\documentclass[11pt]{article}

\usepackage{cmap}
\usepackage[margin=1in]{geometry}
\usepackage[utf8]{inputenc}
\usepackage[T1]{fontenc}
\usepackage{lmodern}
\usepackage{microtype}
\usepackage{amsmath,amssymb}
\usepackage{graphicx}
\usepackage{booktabs}
\usepackage{caption}
\usepackage[section]{placeins}
\usepackage{hyperref}
\usepackage[round]{natbib}
\usepackage{tikz}
\usepackage{authblk}
\usetikzlibrary{arrows.meta,positioning,fit,calc,shapes.geometric,backgrounds,decorations.pathreplacing}

\title{Evaluating Synthetic Skin Lesion Images for Classifier Training: A Safety-Aware Algorithm}

\author{Amindu Kumarasinghe\thanks{\texttt{amindukumarasinghe@gmail.com}}}
\author{Vinh Bui}
\author{Reza Ghanbarzadeh}
\affil{Faculty of Science and Engineering, Southern Cross University, Gold Coast, QLD, Australia}

\date{}

\begin{document}

\maketitle

\begin{abstract}
Skin cancer classifiers need large sets of annotated images, which are scarce, especially for rare lesion
types. AI-generated images can fill the gap, but a generator produces images of widely varying quality, so
selecting which images to retain is important. That choice usually rests on a single quality score,
most often the Fr\'echet
Inception Distance (FID), which has not been tested against what a classifier needs and cannot tell whether
the generator memorizes its training photographs. We test that practice with a tournament of 16 quality
metrics. Each metric alone selects a training subset from a pool of 275{,}472 images generated by a
diffusion model (EDM2), and a classifier trained on each subset scores the metric per class. The tournament
shows that no single metric serves every class: FID computed on Inception features gives the best
classifier for one lesion type and no measurable advantage over random sampling for another, and the
generation setting it prefers is also the one closest to the training images. We turn these per-class
measurements into a safety-aware evaluation algorithm that weights each metric by its measured value in each
class and removes images too close to a real training image. On the held-out ISIC-2018 test set, the
algorithm improves Macro-$F_2$, a recall-weighted measure of classification performance, over the FID
baseline for a Swin-B
classifier by $+0.0136$ to $+0.0164$ across four safety thresholds (Bonferroni-corrected $p<0.02$) and over
random sampling by about $+0.03$, though not for ResNet-50. Tightening the threshold improves a
memorization-safety score with no measurable change in utility, but every curated set remains less safe than
random sampling. The results show a modest but statistically significant gain on the primary classifier, and offer a
practical, safety-aware procedure for evaluating and using synthetic data.
\end{abstract}


\section{Introduction}
\label{sec:intro}

Skin cancer is among the most common cancers worldwide, and its most lethal form,
melanoma, kills thousands of patients every year \citep{who2025, siegel2019}. However,
early detection substantially improves survival, and deep learning classifiers trained on
dermoscopic images can now approach expert-level diagnostic performance
\citep{esteva2017}. That performance depends on large, expertly annotated training sets. These are expensive to
assemble, and for the rarest diagnoses they do not exist in quantity \citep{alzubaidi2023}. Patient privacy
regulations such as the General Data Protection Regulation (GDPR) and the Health Insurance Portability and Accountability Act (HIPAA) add a further constraint, restricting how dermoscopic
images can be shared or pooled across institutions even when they do exist
\citep{giouroukou2025}. On the International Skin Imaging Collaboration (ISIC) 2018 dataset used throughout this paper
\citep{codella2019, tschandl2018}, the seven diagnostic classes, melanoma (MEL), nevus
(NV), basal cell carcinoma (BCC), actinic keratoses and intraepithelial carcinoma (AKIEC), benign
keratosis (BKL),
dermatofibroma (DF), and vascular lesion (VASC), range from 5,364 training images
for the majority class (NV) to 92 for the rarest (DF). This imbalance reflects how rare each diagnosis is in the clinic, so the rarest classes stay small
in any realistic collection of clinical images.

Synthetic image generation is the field's response to that scarcity. Generative
Adversarial Networks \citep{goodfellow2014} and, more recently, diffusion models
\citep{ho2020, rombach2022} can produce class-conditioned dermoscopic images in the
hundreds of thousands, at a cost that has nothing to do with how rare the underlying
condition is. This removes the original scarcity problem and raises a new one: with a
pool that large, how do we tell which images actually help a classifier detect skin
cancer, and which hurt it? That is the curation problem this paper addresses.

Fr\'echet Inception Distance (FID) is one of the most commonly used metrics for
evaluating generative models \citep{heusel2017, betzalel2024}. It measures distance in
the feature space of an Inception-v3 network trained to classify natural photographs, not
dermoscopic images \citep{kynkaanniemi2023}; computed this way it is called
FID-Inception, the label used throughout this paper. The standard practice is nonetheless to
select the single generation checkpoint with the best FID-Inception score and sample
whatever synthetic images are needed from it alone \citep{wu2026}. This is a rule of thumb, not a
measured choice: nothing in a metric built around natural-image classification guarantees
that the checkpoint it prefers on dermoscopic data is also the one that produces a better
classifier, and a recent study in a comparable medical imaging setting reports a
similar mismatch between FID and downstream performance \citep{wu2026}. We treat this gap between what a
metric measures and what a classifier actually needs as an empirical
question rather than an assumption, and we answer it directly, with a procedure we call
the metric tournament. We first used a diffusion model (EDM2) to generate a 275,472-image
synthetic pool spanning ISIC-2018's seven classes. We then selected 16 candidate metrics,
which fall into two families:
instance-level metrics, which score one image on its own and range from pixel-level
measures like peak signal-to-noise ratio (PSNR) to perceptual and semantic ones like learned perceptual image patch similarity (LPIPS), feature similarity index (FSIM), and contrastive language-image pretraining (CLIP) similarity,
and distributional metrics such as FID-Inception, Kernel Inception Distance (KID), and
Sliced Wasserstein Distance (SWD), which compare a whole set of generated images against real
data and cannot judge a single image in isolation. For each metric in turn, we used it
alone to select a training subset from the pool that follows the real class proportions,
trained Swin-B, a transformer-based image classifier and our primary architecture,
on that subset (with ResNet-50 as a secondary check), and measured its per-class $F_2$ score, which weights recall
over precision, against an identically trained classifier given an equally sized,
randomly sampled subset instead. We call this the \emph{utility} of a
metric: its actual downstream performance.

The tournament's result is that no single metric serves every class, and the practice of
picking a single metric across all classes does not hold up. FID-Inception, for
instance, produces the best classifier on one class (VASC) and one no better than
random sampling on another (DF); the full per-class breakdown across all 16 metrics is in
Section~\ref{sec:tournament}. Curating with a single metric helps some classes and is associated with lower downstream
performance for others, and there is no way to know which without measuring it directly.

This paper's algorithm is built on that measurement. It converts the
tournament's per-class measurements into per-class
sampling weights. Distributional metrics describe a whole set of images, so the algorithm
adds a mechanism that lets them shape individual sampling decisions as well.
Finally, a generative
model can sometimes reproduce a specific training image rather than a genuinely new one,
a privacy risk this paper calls memorization; the algorithm gates the resulting pool with
a memorization-safety threshold, calibrated against pairs of real photographs of the same
lesion (ISIC lesion identifiers), that removes images whose
feature-space similarity to a real training image reaches the threshold. This threshold is a tunable lever, and the confirmatory benchmark below tests it
at four calibration-grounded values.

We then validate the algorithm the way it would actually be used, as a real-data
augmentation method, across one further axis beyond the utility already measured by the
tournament: safety, the memorization risk just described. Evaluation is on a
strictly held-out real ISIC-2018 test set that never enters curation, model selection, or
the metric tournament just described, with the classifier's initialization and training
procedure held fixed across every run. Across the four safety threshold values tested
below, it improves macro-averaged $F_2$ over the FID-Inception baseline by $+0.0136$ to
$+0.0164$, with Welch's $t$-test and Bonferroni correction, $p<0.02$ at every one (Section~\ref{sec:benchmark}). Safety depends on the threshold: tightening it raises the
algorithm's memorization-safety score above FID-Inception's while Macro-$F_2$ stays flat, but every
curated set, the algorithm's included, remains less safe than random sampling. The full breakdown is in Section~\ref{sec:safety-threshold-effect}.

This paper asks two primary and two secondary questions. The primary questions are (1)~how well do
current metrics work as selection signals for skin-lesion classifiers, and (2)~can an algorithm built on
these measurements exceed the current standard, FID-based selection, on both utility and safety. The
secondary questions are (3)~does the feature backbone behind a metric, Inception or DINOv2, matter, and
(4)~do the gains of the algorithm hold when it curates a larger set.

The contributions of this paper are: (1)~an empirical metric tournament that evaluates 16 current metrics as selection signals for
skin-lesion classifiers and shows that no single metric serves every class; (2)~a safety-aware evaluation algorithm built on
the results of the tournament; (3)~a confirmatory
benchmark on a real, held-out clinical test set, where it outperforms the
industry-standard FID-Inception baseline on utility and, at the two tighter thresholds, on
safety as well; and (4)~a public release of the
275,472-image synthetic
pool with per-image scores for 14 metrics and per-configuration scores for the four distributional
metrics, together with open-source code. These findings bear on how synthetic images should be
selected for rare-lesion classifiers, where a missed diagnosis is costly: the choice of selection
metric should be validated per class on held-out real data (Section~\ref{sec:discussion-clinical}).

Following this introduction, Section~\ref{sec:related-work} situates this work against
prior synthetic augmentation, generative evaluation, and memorization literature.
Section~\ref{sec:methods} describes the full pipeline: the synthetic pool, the metric tournament,
the evaluation algorithm, the safety measure and the benchmark protocol.
Section~\ref{sec:results} reports the tournament (question 1), the feature backbone of the metrics
(question 3), the development evidence for each of the algorithm's design choices, the confirmatory
benchmark with its safety analysis and its dose response (questions 2 and 4), and a repetition of the
pipeline with ResNet-50 to test how far the findings generalize.
Section~\ref{sec:dataset-release} describes the released dataset.
Sections~\ref{sec:discussion} to~\ref{sec:conclusion} discuss what the algorithm does and
does not establish, its limitations, and its conclusions.

\section{Related Work}
\label{sec:related-work}

Synthesis of dermoscopic images with generative adversarial networks (GANs) and diffusion models has been studied extensively, but
its effect on downstream classifier performance is inconsistent across studies using
near-identical setups. Bissoto et al.\ tested four GAN architectures under multiple
sampling strategies and real-to-synthetic ratios and concluded that ``reliably
transforming synthetic images into reliable performance gains is far from obvious'':
augmentation helped on out-of-distribution test sets but not on standard in-distribution
dermoscopic classification, and which conditions transfer remained an open question
\citep{bissoto2021}. The same pattern recurs elsewhere. Miko{\l}ajczyk et al.\ found that
an unconditional GAN's synthetic images nearly matched real-only training performance
while a conditional GAN's images degraded it, despite sharing the same underlying
architecture family \citep{mikolajczyk2022}. An earlier synthesis study by Bissoto et
al.\ found that only one of eight tested real-synthetic training compositions improved
area under the receiver operating characteristic curve (ROC-AUC) over a real-only baseline, with several compositions performing worse
\citep{bissoto2018}. Ding et al.\ found that only their own proposed conditional GAN's
synthetic images improved downstream ROC-AUC when added to real data, while three
alternative generators' outputs did not \citep{ding2021}. Curation, not generation, is
also what drives benefit when clinicians are put in the loop directly: Nagi et al.\ had
pediatric dermatologists approve or reject individual synthetic images before
augmentation, and found that training on the clinician-approved subset alone beat both a
real-only baseline and augmentation with the full, unfiltered synthetic pool, with the
largest gain in an under-represented dark-skin-tone subgroup (though the authors flag
that specific subgroup result as exploratory, given only seven such patients in the
comparison); they state plainly that ``visually plausible synthetic images may still
contain clinically implausible features... not adequately captured by automatic
image-quality metrics'' \citep{nagi2026}. The same inconsistency shows up with diffusion
models: Wang et al.\ found diffusion-generated skin-disease images were frequently
medically inaccurate and degraded downstream diagnostic performance until corrected
against expert feedback, after which accuracy improved by 9 to 14 percentage points
depending on the data regime \citep{wang2025}, while Shentu et al.\ report a more modest
+2.32 F1 gain from a different diffusion-based pipeline \citep{shentu2026}. The same
basic idea, generation followed by augmentation, produces effects ranging from harmful to
strongly beneficial depending on what stands between the generator and the training set.
Across these studies, the recurring pattern is that generation quality alone does
not predict downstream benefit:
which images are added, not only how they were made, affects whether augmentation
helps or hurts. That is the curation problem this paper addresses.

When studies of synthetic augmentation in dermatology report a quantitative image-quality metric at all, they typically use
one, chosen by convention rather than by what it predicts about the downstream task.
Alonso et al.\ report only FID \citep{alonso2025}; Teodoro
et al.\ track FID alone across training to certify a generator as sufficient for their
pipeline \citep{teodoro2022}; Rezk et al.\ use two instance-level metrics, the blind/referenceless image spatial quality evaluator (BRISQUE) and
the structural similarity index (SSIM), each assigned to a separate purpose (realism and disease presentation
respectively) rather than combined into one measured score \citep{rezk2022}. Luschi et
al.\ tested this directly. Using a StyleGAN2 model already selected via FID, KID,
precision, and recall in a companion study, they measured per-image LPIPS for 25
synthetic melanoma images and correlated it against the diagnostic accuracy of 17
dermatologists assessing those same images, finding no significant relationship
(Spearman's $\rho=-0.36$ to $-0.24$, $p>0.05$); the dermatologists themselves classified
real from synthetic images at only 64\% accuracy. They conclude that ``common
quantitative metrics are not able to capture context-specific aspects of realism''
\citep{luschi2025}. A direct architecture comparison makes the same point more starkly:
Pritam and Jain compared StyleGAN2-ADA against a denoising diffusion probabilistic model (DDPM) across FID, KID, Inception Score (IS),
and downstream classification on over 14,000 dermoscopic images, and found that the
architecture with the better distributional scores did not simply produce the better
downstream classifier, with augmentation gains ranging from 8\% to 15\% F1 depending on
architecture and preprocessing choices that barely moved the generative metrics
themselves \citep{pritam2025}. At the opposite extreme, some evaluations use no
quantitative metric at all: Pielage et al.\ validated their diffusion-generated lesions
through a dermatologist user study alone, with no FID, KID, or downstream classifier
reported \citep{pielage2025}. Bissoto et al.\ question the
premise directly: FID and the IS rely on ImageNet-trained networks, and are
``questionable for applications where classes are very different, such as medical
applications'' \citep{bissoto2021}. The same disconnect has been reported outside dermoscopy. Ravuri and Vinyals scored class-conditional
generative models by the accuracy of a classifier trained only on their images and tested on real data,
and found large deficits that FID and similar scores did not reveal \citep{Ravuri2019}. Skandarani et
al.\ report that GANs that look realistic by FID standards still fail to reproduce the richness of medical
datasets for downstream segmentation \citep{Skandarani2023}. Jayasumana et al.\ attribute part of FID's
unreliability to its normality assumption, its Inception features and its sample complexity, and propose
CLIP maximum mean discrepancy (CMMD), which uses CLIP embeddings and the maximum mean discrepancy (MMD) \citep{jayasumana2024}. Stein et al.\
find that no existing metric correlates strongly with human judgment of realism, that over-reliance on
Inception features is one reason FID misses the realism of diffusion models, and that DINOv2 features allow
a richer evaluation \citep{stein2023}. In
medical imaging, Woodland et al.\ find that ImageNet-based feature extractors rank generators more
consistently with expert judgment than extractors trained on medical images \citep{woodland2024}. This paper treats these critiques as a reason to
measure, rather than assume, each metric's downstream value: the 16-metric portfolio in
Section~\ref{sec:tournament-metric-portfolio} spans exactly the instance-level and distributional families
used piecemeal above, each scored against classifier performance.

Selecting training examples has a long history in machine learning, for example core-set selection
\citep{sener2018}, forgetting events \citep{toneva2019} and importance scores computed early in training
\citep{paul2021}. These methods rank real training examples using a model trained on them. In our setting the
candidates are synthetic and each score is computed from the image and its features, with the classifier
entering only through the tournament that sets the weights. Filtering a synthetic pool has also been tried
in dermoscopy. Islam et al.\ remove synthetic skin-lesion
images with less discriminative features than the real training images, judged by cosine similarity
\citep{Islam2024}, a filter aimed at class separability. The safety filter of this paper targets the
opposite end, images that are too close to a real training image.

Generative models can also memorize and reproduce individual training examples rather
than generating novel ones. Carlini et al.\ extracted over a thousand training
images from state-of-the-art diffusion models, ``ranging from photographs of individual
people to trademarked company logos,'' and found diffusion models ``much less private
than prior generative models such as GANs'' \citep{carlini2023}. In medical imaging
specifically, this has been demonstrated quantitatively: Akbar
et al.\ found diffusion models more likely to memorize training images than StyleGAN when
synthesizing brain magnetic resonance imaging (MRI) and chest X-ray images, particularly on small datasets
\citep{akbar2023}, and Dar et al.\ detected memorization in approximately 37\% of patient
scans across computed tomography (CT), MRI, and X-ray diffusion models, concluding that ``a synthesized copy
can even be traced back to the original patient, leading to patient re-identification''
\citep{dar2025}. Meehan et al.\ formalized this as data-copying, in which a generative model outputs training samples or
small variations of them, and proposed a three-sample test to detect it \citep{Meehan2020}. Stein et al.\ likewise find that current
generative metrics do not properly detect memorization \citep{stein2023}.
No equivalent quantitative study yet exists for dermoscopic or
skin-lesion generation specifically; the closest is Bissoto et al.'s own observation, in
the context of GAN-based anonymization, that ``we need further studies on how much the
GAN `remembers' each original training sample'' \citep{bissoto2021}, a flagged concern
rather than a measured one. This paper does not assume dermoscopy inherits the exact
memorization rates documented in radiology; it treats the absence of a
dermatology-specific measurement as a reason for a precautionary, ground-truth-calibrated
safety filter (Section~\ref{sec:algorithm}), not as evidence that no filter is needed.

Taken together, these three threads point to the same missing piece. Generation quality
alone does not settle downstream benefit and the choice of training images matters, but the existing
literature provides limited guidance on image selection. The metrics available for making that choice
are used narrowly and are
rarely checked against the task they are meant to serve. And the resulting pool's
memorization risk is, in dermatology specifically, unmeasured rather than shown to be
absent. The safety-aware evaluation algorithm introduced in Section~\ref{sec:algorithm} treats each
of these as something to measure rather than assume: curation weights are set
from metrics whose downstream value has actually been tested (Section~\ref{sec:tournament}),
and the resulting pool is gated by a safety threshold calibrated against real ground
truth, precisely because the measurement gap identified above has not been closed
elsewhere.

\section{Methods}
\label{sec:methods}

Synthetic image quality is usually judged with a single metric, most often FID \citep{alonso2025, teodoro2022}, on the
assumption that a metric that scores a generator well also selects images that help a
classifier \citep{wu2026}. The method begins by testing that assumption directly. Each of 16 metrics is used on its own
as a curation criterion, a classifier is trained on the images that metric selects, and the
metric is judged by how well that classifier performs. The result is a measurement of
the downstream utility of each metric. Section~\ref{sec:algorithm} converts these measurements
into curation weights.

This section describes the full pipeline. Sections~\ref{sec:methods-real-data} to~\ref{sec:tournament-tournament-protocol} define the real data, the synthetic pool, the metrics and the tournament protocol. Section~\ref{sec:algorithm} defines the evaluation algorithm, Section~\ref{sec:methods-fsms} the safety measure and Section~\ref{sec:benchmark-protocol} the benchmark protocol. The results follow in Section~\ref{sec:results}.

\subsection{Real Data Split}
\label{sec:methods-real-data}

We split the 10,015-image ISIC-2018 training archive \citep{codella2019, tschandl2018} into a
class-stratified 80/20 partition: 8,012 images for training and 2,003 for validation. The training
partition trains the generator and is the real reference set of every reference-based metric. The
validation partition scores the classifiers of the tournament. The official ISIC-2018 test set is held
out for the benchmark of Section~\ref{sec:benchmark}.

\subsection{Synthetic Pool}
\label{sec:tournament-synthetic-pool}

The pool contains 275,472 images generated by class-conditional EDM2 models \citep{Karras2024}
trained on the 8,012-image real training partition. Images were sampled from 16 generation
configurations, formed by four training checkpoints (3k, 8k, 15k and 25k kimg) crossed with four
classifier-free guidance (CFG) scales (1.0, 1.5, 2.0 and 3.0). The checkpoints sample the generator from early (3k kimg) to late (25k kimg) in training, so that the pool contains images of varying quality. The CFG scales vary the fidelity-diversity trade-off of the pool. A scale of
1.0 is unguided, 1.5 and 2.0 lie within the range of CFG scales (1.2 to 2.0) used for the published
EDM2 ImageNet-512 models, and 3.0 is a stronger setting beyond it \citep{Karras2024}. Each configuration
contributes 17,217
images: 1,786 for each of the six minority classes and 6,501 for NV (Table~\ref{tab:pool-composition}). The additional NV images
ensure that any single configuration can supply a complete curated set with the real class proportions on its own,
which the distributional arms below require.

\begin{table}[!ht]
\centering
\caption{Composition of the synthetic pool: images per generation configuration (training checkpoint
in kimg by CFG scale) and class.}
\label{tab:pool-composition}
\footnotesize
\setlength{\tabcolsep}{4pt}
\begin{tabular}{@{}llrrrrrrrr@{}}
\toprule
Checkpoint & CFG & AKIEC & BCC & BKL & DF & MEL & NV & VASC & Total \\
\midrule
3k & 1.0 & 1,786 & 1,786 & 1,786 & 1,786 & 1,786 & 6,501 & 1,786 & 17,217 \\
 & 1.5 & 1,786 & 1,786 & 1,786 & 1,786 & 1,786 & 6,501 & 1,786 & 17,217 \\
 & 2.0 & 1,786 & 1,786 & 1,786 & 1,786 & 1,786 & 6,501 & 1,786 & 17,217 \\
 & 3.0 & 1,786 & 1,786 & 1,786 & 1,786 & 1,786 & 6,501 & 1,786 & 17,217 \\
\midrule
8k & 1.0 & 1,786 & 1,786 & 1,786 & 1,786 & 1,786 & 6,501 & 1,786 & 17,217 \\
 & 1.5 & 1,786 & 1,786 & 1,786 & 1,786 & 1,786 & 6,501 & 1,786 & 17,217 \\
 & 2.0 & 1,786 & 1,786 & 1,786 & 1,786 & 1,786 & 6,501 & 1,786 & 17,217 \\
 & 3.0 & 1,786 & 1,786 & 1,786 & 1,786 & 1,786 & 6,501 & 1,786 & 17,217 \\
\midrule
15k & 1.0 & 1,786 & 1,786 & 1,786 & 1,786 & 1,786 & 6,501 & 1,786 & 17,217 \\
 & 1.5 & 1,786 & 1,786 & 1,786 & 1,786 & 1,786 & 6,501 & 1,786 & 17,217 \\
 & 2.0 & 1,786 & 1,786 & 1,786 & 1,786 & 1,786 & 6,501 & 1,786 & 17,217 \\
 & 3.0 & 1,786 & 1,786 & 1,786 & 1,786 & 1,786 & 6,501 & 1,786 & 17,217 \\
\midrule
25k & 1.0 & 1,786 & 1,786 & 1,786 & 1,786 & 1,786 & 6,501 & 1,786 & 17,217 \\
 & 1.5 & 1,786 & 1,786 & 1,786 & 1,786 & 1,786 & 6,501 & 1,786 & 17,217 \\
 & 2.0 & 1,786 & 1,786 & 1,786 & 1,786 & 1,786 & 6,501 & 1,786 & 17,217 \\
 & 3.0 & 1,786 & 1,786 & 1,786 & 1,786 & 1,786 & 6,501 & 1,786 & 17,217 \\
\midrule
All 16 & & 28,576 & 28,576 & 28,576 & 28,576 & 28,576 & 104,016 & 28,576 & 275,472 \\
\bottomrule

\end{tabular}
\end{table}

\subsection{Metric Portfolio}
\label{sec:tournament-metric-portfolio}

\begin{figure}[tbp]
\centering
\resizebox{\linewidth}{!}{%
\definecolor{ink}{HTML}{2B2B2B}\definecolor{acc}{HTML}{2F5D8A}\definecolor{accl}{HTML}{E6EEF6}
\definecolor{panel}{HTML}{F4F4F2}\definecolor{outc}{HTML}{FFF4DC}
\begin{tikzpicture}[font=\footnotesize, color=ink,
  n/.style={draw=ink, fill=white, rounded corners=2pt, align=center, line width=.6pt, minimum height=1.05cm, text width=3.3cm, inner sep=3pt},
  g/.style={n, text width=4.9cm, minimum height=1.55cm, fill=accl, draw=acc},
  o/.style={n, fill=outc, text width=3.1cm, minimum height=1.35cm},
  a/.style={-{Stealth[length=5pt]}, line width=.8pt, color=acc},
  pan/.style={fill=panel, draw=none, rounded corners=4pt},
  lab/.style={font=\footnotesize\bfseries, color=acc, anchor=south west}]
\node[n] (real) at (0,1.5) {Real training images\\8,012};
\node[n] (syn) at (0,-1.5) {Synthetic pool\\275,472 in 16 configurations};
\node[n] (pre) at (4.4,0) {Standardize to\\$512\times512$ and\\extract features};
\node[n] (dino) at (9.0,2.4) {DINOv2-base CLS\\768-d};
\node[n] (incv) at (9.0,-2.4) {InceptionV3\\2,048-d and probabilities};
\node[n] (faiss) at (9.0,0.8) {Exact FAISS search\\(L2-normalized DINOv2)};
\node[n] (knn) at (9.0,-0.8) {Neighbor map\\nearest real image};
\node[g] (dist) at (14.3,2.775) {\textbf{Distributional}\\FID-Inception, FID-DINOv2,\\KID, SWD\\IS contribution (per image)};
\node[g] (per) at (14.3,0.925) {\textbf{Perceptual}\\PSNR, SSIM, FSIM, LPIPS\\BRISQUE};
\node[g] (sem) at (14.3,-0.925) {\textbf{Semantic}\\CLIP prompts, FS-CS\\classifier confidence};
\node[g] (geo) at (14.3,-2.775) {\textbf{Geometric}\\Realism, Diversity};
\node[o] (ckpt) at (20.1,2.775) {Checkpoint results\\(per configuration)};
\node[o] (meta) at (20.1,-0.925) {Metadata file\\(per image)};
\begin{scope}[on background layer]
\node[pan, fit=(real), inner sep=.3cm] (p1) {};
\node[pan, fit=(syn), inner sep=.3cm] (p1b) {};
\node[pan, fit=(dino)(incv)(faiss)(knn), inner sep=.3cm] (p2) {};
\node[pan, fit=(dist)(geo), inner sep=.3cm] (p3) {};
\end{scope}
\node[lab] at (p1.north west) {Real images};
\node[lab] at (p1b.north west) {Synthetic images};
\node[lab] at (p2.north west) {Cached features};
\node[lab] at (p3.north west) {Metric registry};
\draw[a] (real.east) -- ++(.6,0) |- ($(pre.west)+(0,.3)$);
\draw[a] (syn.east) -- ++(.6,0) |- ($(pre.west)+(0,-.3)$);
\draw[a] ($(pre.east)+(0,.3)$) -- ++(.6,0) |- (dino.west);
\draw[a] ($(pre.east)+(0,-.3)$) -- ++(.6,0) |- (incv.west);
\draw[a] (dino.south) -- (faiss.north);
\draw[a] (faiss.south) -- (knn.north);
\draw[a] (p2.east |- pre.east) -- (p3.west |- pre.east);
\draw[a] (dist.east) -- (ckpt.west);
\draw[line width=.8pt, color=acc] (per.east) -- (17.5,0.925) -- (17.5,-2.775);
\draw[line width=.8pt, color=acc] (sem.east) -- (17.5,-0.925);
\draw[line width=.8pt, color=acc] (geo.east) -- (17.5,-2.775);
\draw[a] (17.5,-0.925) -- (meta.west);
\end{tikzpicture}
}
\caption{The metric pipeline. Real and synthetic images are standardized to $512\times512$ pixels,
and DINOv2 and InceptionV3 features are extracted once and cached. An exact FAISS search in DINOv2
space writes the nearest real image of every synthetic image to a neighbor map. The registry runs
four metric groups over the cached features and the map. The four per-configuration distributional
scores are written to the checkpoint results file, and all per-image scores to the metadata file.
Details of every metric are in Appendix~\ref{app:metrics}.}
\label{fig:metric-pipeline}
\end{figure}
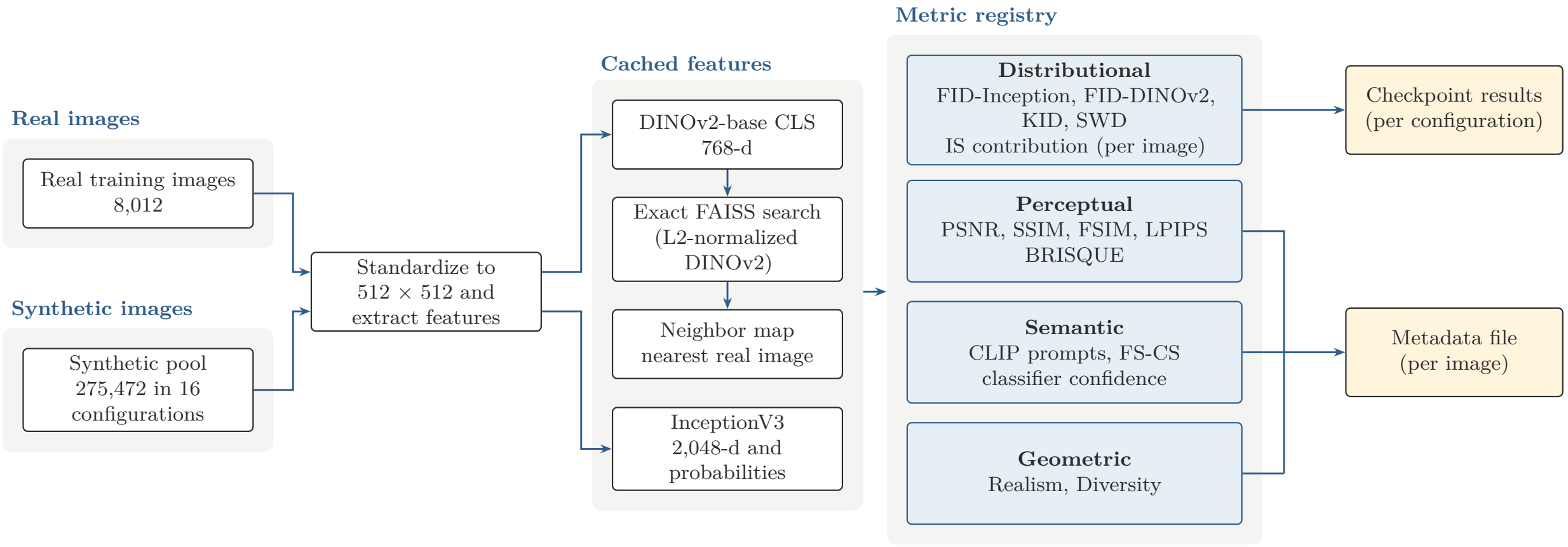

\paragraph{Metrics.}
Sixteen metrics were evaluated (Table~\ref{tab:portfolio}). Twelve score each image
individually (instance metrics). Four score a whole set of images against the real training
images (distributional metrics). Figure~\ref{fig:metric-pipeline}
shows how the scores are produced.

\paragraph{Features.}
All images are first brought to $512 \times 512$ pixels, and
DINOv2 \citep{Oquab2023} and InceptionV3 features are extracted once and cached. The DINOv2 feature of
an image is the 768-dimensional class token (CLS) embedding of DINOv2-base, which summarizes the whole
image. DINOv2 is the main feature space and underlies every feature-based metric except the Inception
Score and FID-Inception.
InceptionV3 is included for comparison, because FID computed on its features is the
standard image-quality measure in the generative modeling literature.

\paragraph{Per-image scores.}
The full-reference metrics (PSNR, SSIM, FSIM and LPIPS) compare an image with a reference. The generator is
not paired with any real image, so the reference is the nearest real training image in DINOv2 space, found
with an exact nearest-neighbor search in FAISS \citep{Johnson2017faiss}. BRISQUE \citep{Mittal2012} is a
no-reference score of natural-image statistics. Feature-space cosine similarity (FS-CS) is a metric we
define in this paper. For an image generated for class $c$ it is the cosine similarity between the image's
DINOv2 feature and the class centroid, the mean DINOv2 feature of the real training images of class $c$.
A high value means that the image lies close to the center of its real class in a feature space that
carries semantic content. The CLIP scores \citep{Radford2021} are the similarities of an image to two text
prompts, one describing a clinical photograph and one a computer-generated image, and the log-odds of the
two. The IS is decomposed into its per-image Kullback-Leibler (KL) divergence term, which gives one value per
image. Realism and Diversity are adapted from the manifold-based precision and recall of
\citet{Kynkaeaenniemi2019} so that they return one continuous score per image. Realism is the distance to
the third nearest real image and Diversity the distance to the third nearest other synthetic image, both in
DINOv2 space and found with the same exact search.

\paragraph{Distributional metrics.}
The distributional metrics compare the whole image set of a generation configuration (17,217 images) with
the 8,012 real training images, so each configuration receives one score per metric, and lower is better.
FID \citep{heusel2017} is the Fr\'echet distance between Gaussians fitted to the two sets of features. It is
computed on InceptionV3 features (FID-Inception) and on DINOv2 features (FID-DINOv2). KID
\citep{Binkowski2018} is the squared MMD between the two sets with a cubic
polynomial kernel, and SWD \citep{Bonneel2015} is the sliced Wasserstein distance, the mean absolute
difference between the sorted one-dimensional projections of the two sets onto random directions. KID and
SWD are computed on DINOv2 features. Appendix~\ref{app:metrics} specifies every metric in detail.

\begin{table}[t]
\centering
\caption{The 16 metrics in the tournament. ``Selects'' gives the direction in which a metric
ranks images (instance metrics) or configurations (distributional metrics).}
\label{tab:portfolio}
\footnotesize
\begin{tabular}{@{}p{2.7cm}p{3.9cm}p{6.2cm}l@{}}
\toprule
Family & Metric & What it scores & Selects \\
\midrule
Distributional & FID-Inception \citep{heusel2017} & Synthetic vs.\ real, Inception features & lowest \\
 & FID-DINOv2 & Synthetic vs.\ real, DINOv2 features & lowest \\
 & KID-DINOv2 \citep{Binkowski2018} & Synthetic vs.\ real, polynomial-kernel MMD & lowest \\
 & SWD-DINOv2 \citep{Bonneel2015} & Synthetic vs.\ real, sliced Wasserstein & lowest \\
\midrule
Instance, full-reference & PSNR & Pixel error to nearest real image & highest \\
 & SSIM \citep{Zhou2004} & Structural similarity to nearest real image & highest \\
 & FSIM \citep{Zhang2011} & Phase-congruency similarity to nearest real image & highest \\
 & LPIPS \citep{zhang2018lpips} & Perceptual distance to nearest real image & lowest \\
Instance, no-reference & BRISQUE \citep{Mittal2012} & Natural-scene statistics deviation & lowest \\
Instance, semantic & FS-CS & DINOv2 cosine similarity to real class centroid & highest \\
 & CLIP sim.\ (cancer) \citep{Radford2021} & Similarity to a clinical-photograph prompt & highest \\
 & CLIP sim.\ (artifact) & Similarity to a computer-generated-image prompt & lowest \\
 & CLIP log-odds & Log ratio of the two CLIP similarities & highest \\
 & IS contribution \citep{Salimans2016} & Per-image KL term of the IS & highest \\
Instance, geometric & Realism & Distance to 3rd nearest real image (DINOv2) & lowest \\
 & Diversity & Distance to 3rd nearest other synthetic image (DINOv2) & highest \\
\bottomrule
\end{tabular}
\end{table}

\subsection{Tournament Protocol}
\label{sec:tournament-tournament-protocol}

\paragraph{Metric isolation.}
Every metric curates a dataset of exactly 8,012 synthetic images, matching the size of the real
training set. Class proportions follow the real training set (AKIEC 3.27\%, BCC 5.13\%,
BKL 10.97\%, DF 1.15\%, MEL 11.11\%, NV 66.95\%, VASC 1.42\%; quotas are rounded down and the remainder goes to NV), so class priors are identical
across arms. This isolates the metric: if the class balance differed between arms, a difference in
classifier performance could come from the balance and not from the metric that selected the images. An instance metric ranks the entire pool within each class and keeps the top
images that fill the class quota. This procedure is deterministic. Therefore, a single dataset
represents the selection produced by each metric. A distributional metric identifies the single generation
configuration with the best score and fills each class quota by uniform random sampling from that
configuration alone. The four metrics selected three configurations: FID-Inception chose
the 25k checkpoint at guidance 1.0, FID-DINOv2 and KID-DINOv2 both chose 25k at 1.5, and SWD-DINOv2
chose 15k at 1.5 (scores of all 16 configurations in Table~\ref{tab:distributional-pool}, Appendix~\ref{app:tournament}). FID-DINOv2 and KID-DINOv2 happened to choose the same configuration and
share seeds, so their curated datasets and results are identical. KID is a different statistic from
FID, and the match is a coincidence of which configuration scored best.

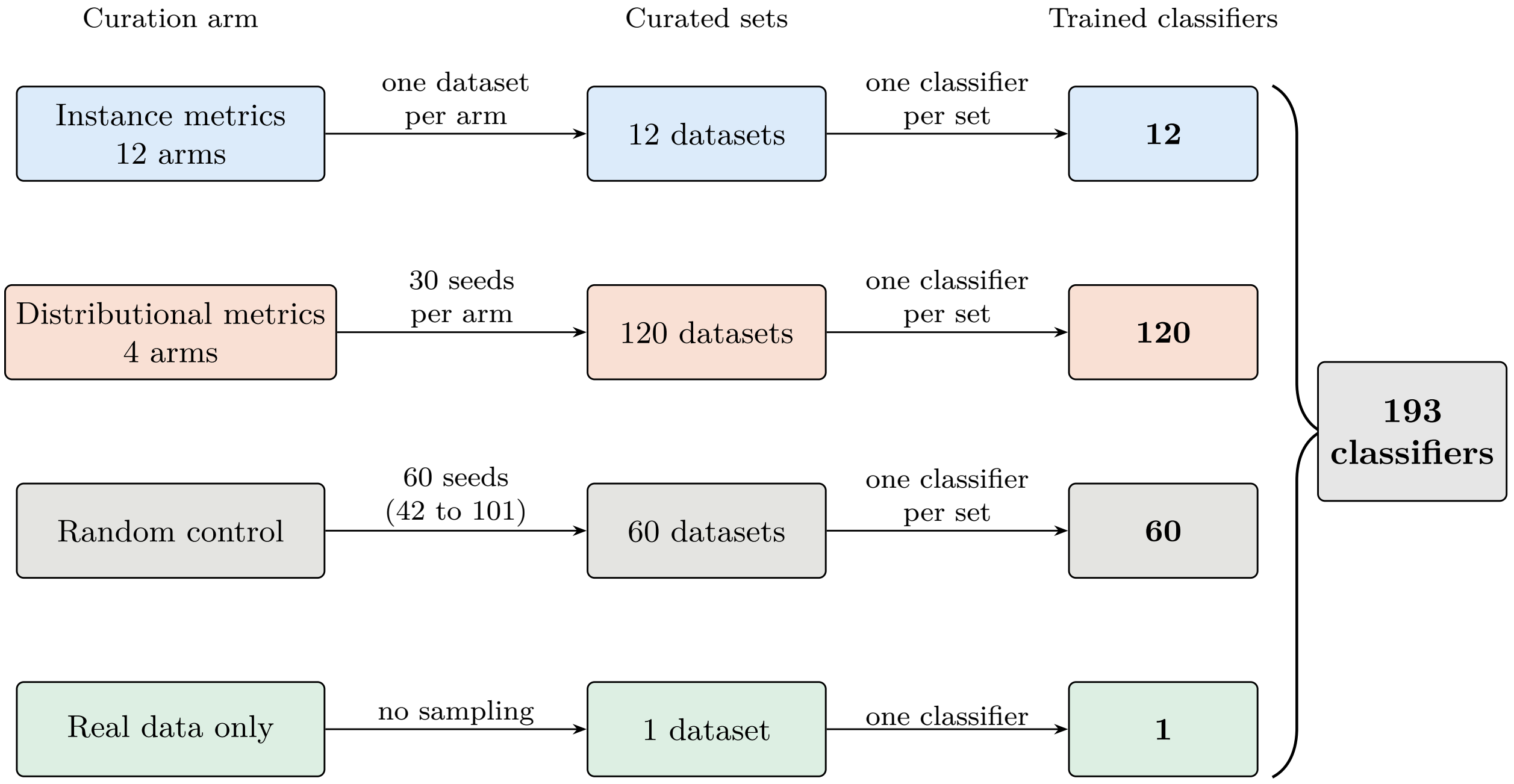
\begin{figure}[tbp]
\centering
\resizebox{\linewidth}{!}{%
\begin{tikzpicture}[
    font=\footnotesize,
    box/.style={draw, rounded corners=2pt, minimum height=0.95cm, align=center, inner sep=3pt, line width=0.5pt},
    src/.style={box, fill=black!6, minimum width=2.5cm},
    arm/.style={box, minimum width=3.1cm},
    ds/.style={box, minimum width=2.4cm},
    res/.style={box, minimum width=1.9cm, font=\footnotesize\bfseries},
    tot/.style={box, fill=black!10, minimum width=1.9cm, minimum height=1.4cm, font=\small\bfseries},
    flow/.style={-{Stealth[length=4pt]}, line width=0.5pt},
    lab/.style={font=\scriptsize, align=center, inner sep=1pt}]
\definecolor{cinst}{HTML}{DCEBFA}
\definecolor{cdist}{HTML}{F9E0D4}
\definecolor{crand}{HTML}{E4E4E1}
\definecolor{creal}{HTML}{DDEFE3}
\node[arm, fill=cinst] (a1) at (0,3.0) {Instance metrics\\12 arms};
\node[arm, fill=cdist] (a2) at (0,1.0) {Distributional metrics\\4 arms};
\node[arm, fill=crand] (a3) at (0,-1.0) {Random control};
\node[arm, fill=creal] (a4) at (0,-3.0) {Real data only};
\node[ds, fill=cinst] (d1) at (5.4,3.0) {12 datasets};
\node[ds, fill=cdist] (d2) at (5.4,1.0) {120 datasets};
\node[ds, fill=crand] (d3) at (5.4,-1.0) {60 datasets};
\node[ds, fill=creal] (d4) at (5.4,-3.0) {1 dataset};
\node[res, fill=cinst] (r1) at (10.0,3.0) {12};
\node[res, fill=cdist] (r2) at (10.0,1.0) {120};
\node[res, fill=crand] (r3) at (10.0,-1.0) {60};
\node[res, fill=creal] (r4) at (10.0,-3.0) {1};
\draw[flow] (a1) -- node[lab, above] {one dataset\\per arm} (d1);
\draw[flow] (a2) -- node[lab, above] {30 seeds\\per arm} (d2);
\draw[flow] (a3) -- node[lab, above] {60 seeds\\(42 to 101)} (d3);
\draw[flow] (a4) -- node[lab, above] {no sampling} (d4);
\draw[flow] (d1) -- node[lab, above] {one classifier\\per set} (r1);
\draw[flow] (d2) -- node[lab, above] {one classifier\\per set} (r2);
\draw[flow] (d3) -- node[lab, above] {one classifier\\per set} (r3);
\draw[flow] (d4) -- node[lab, above] {one classifier} (r4);
\draw[decorate, decoration={brace, amplitude=14pt, raise=4pt}, line width=0.8pt] (r1.north east) -- (r4.south east) node[midway, xshift=1.55cm, tot] {193\\classifiers};
\node[lab, above=0.55cm of a1] {Curation arm};
\node[lab, above=0.55cm of d1] {Curated sets};
\node[lab, above=0.55cm of r1] {Trained classifiers};
\end{tikzpicture}}
\caption{The 193 Swin-B classifiers of the tournament. The 12 instance arms and the 4 distributional
arms come from the synthetic pool. The 30 seeds of each distributional arm and the 60 seeds of the
random control give the curated sets, and one classifier is trained on each set.}
\label{fig:tournament-runs}
\end{figure}

\paragraph{Seeds and baselines.}
Random sampling within the chosen configuration makes the distributional arms stochastic, so each
was curated and trained 30 times (seeds 42 to 71). Each distributional arm draws from its own
selected configuration's pool while the random control draws from the whole pool, so the two are
treated as independent samples (Welch's $t$-test). The random control also anchors the
metric weights of Section~\ref{sec:algorithm}, so it was run with 60 seeds (42 to 101) to estimate
that baseline more precisely, and all 60 runs serve as its comparison with each arm. Instance arms were run once, because their selection is deterministic. A real-data-only classifier trained on
the 8,012 real images is the reference. In total the tournament comprises 193 trained
classifiers (12 instance, 120 distributional, 60 random control, 1 real-only), as
Figure~\ref{fig:tournament-runs} shows.

\paragraph{Classifiers.}
Every arm is evaluated with a Swin-B classifier \citep{Liu2021} that starts from ImageNet-pretrained
weights with the backbone frozen, so only the final classification layer is trained. Freezing the
backbone fixes the feature space for every arm. This extends the linear probe of \citet{alain2016understanding}, 
which reads a model's representations without altering them, to a measurement of the quality of a curated 
dataset. A difference between two arms is then attributable to the images
they contain, whereas fine-tuning also moves the backbone toward each training set
\citep{kumar2022fine}. The optimizer is
AdamW with a learning rate of $10^{-3}$ and a batch size of 32, for at most 50 epochs with early
stopping (patience 10 on validation accuracy), and the inputs are random-resized-cropped to 224
pixels and flipped horizontally. Weight initialization and data-loader seed are fixed at 42 in every
run, so the only quantity that differs between repetitions of an arm is which synthetic images were
selected. Training uses synthetic images only. Section~\ref{sec:resnet} repeats the pipeline with ResNet-50 \citep{He2016} under identical
settings.

\paragraph{Evaluation.}
Each classifier is scored on the real validation split of 2,003 images
(Section~\ref{sec:methods-real-data}). The official ISIC-2018 test set is not used
anywhere in the tournament. It is used only in the confirmatory benchmark of
Section~\ref{sec:benchmark}, which evaluates the algorithm and its baselines. The score is $F_\beta = (1+\beta^2)PR/(\beta^2 P + R)$ for precision $P$ and recall $R$,
with $\beta = 2$, so that $F_2 = 5PR/(4P + R)$. This weights recall more heavily than precision because a missed lesion costs more than a false
alarm. It is computed per class and averaged over the seven classes to give Macro-$F_2$.

The same validation split also selects the early-stopping epoch, so absolute scores are optimistic, but the procedure is identical for every arm and favors none.
Absolute performance is evaluated on the held-out test set of Section~\ref{sec:benchmark}. The scores do differ in how much replication stands behind them: a distributional
arm is a mean over 30 draws, whereas an instance arm is one deterministic dataset with one score, so its noise cannot be estimated.

\subsection{The Evaluation Algorithm}

\label{sec:algorithm}

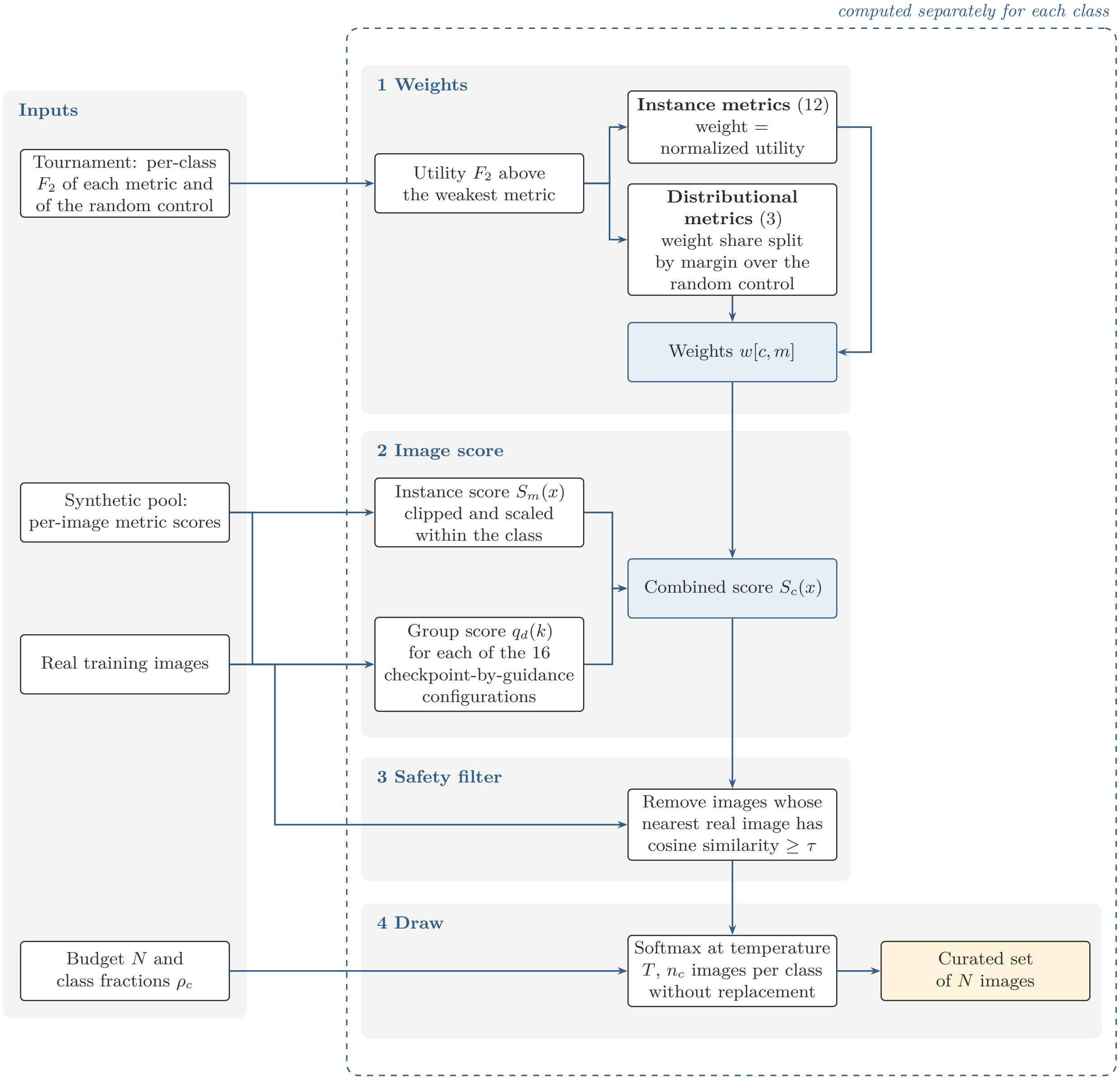
\begin{figure}[tbp]
\centering
\resizebox{\linewidth}{!}{%
\definecolor{ink}{HTML}{2B2B2B}\definecolor{acc}{HTML}{2F5D8A}\definecolor{accl}{HTML}{E6EEF6}
\definecolor{panel}{HTML}{F4F4F2}\definecolor{outc}{HTML}{FFF4DC}
\hyphenpenalty=10000 \exhyphenpenalty=10000 \begin{tikzpicture}[font=\footnotesize, color=ink,
  n/.style={draw=ink, fill=white, rounded corners=2pt, align=center, line width=.6pt, minimum height=1.05cm, text width=3.5cm, inner sep=3pt},
  k/.style={n, fill=accl, draw=acc},
  o/.style={n, fill=outc},
  a/.style={-{Stealth[length=5pt]}, line width=.8pt, color=acc},
  pan/.style={fill=panel, draw=none, rounded corners=4pt},
  lab/.style={font=\footnotesize\bfseries, color=acc, anchor=south west}]
\node[n] (tour) at (0,6.2) {Tournament: per-class $F_2$ of each metric and of the random control};
\node[n] (pool) at (0,0.35) {Synthetic pool: per-image metric scores};
\node[n] (real) at (0,-2.35) {Real training images};
\node[n] (bud) at (0,-7.8) {Budget $N$ and class fractions $\rho_c$};
\node[n] (u) at (6.3,6.2) {Utility $F_2$ above the weakest metric};
\node[n] (wi) at (10.8,7.2) {\textbf{Instance metrics} (12)\\weight $=$ normalized utility};
\node[n] (wd) at (10.8,5.2) {\textbf{Distributional metrics} (3)\\weight share split by margin over the random control};
\node[k] (w) at (10.8,3.2) {Weights $w[c,m]$};
\node[n] (si) at (6.3,0.35) {Instance score $S_m(x)$ clipped and scaled within the class};
\node[n] (sg) at (6.3,-2.35) {Group score $q_d(k)$ for each of the 16 checkpoint-by-guidance configurations};
\node[k] (sc) at (10.8,-1.0) {Combined score $S_c(x)$};
\node[n] (fl) at (10.8,-5.2) {Remove images whose nearest real image has cosine similarity $\geq\tau$};
\node[n] (dr) at (10.8,-7.8) {Softmax at temperature $T$, $n_c$ images per class without replacement};
\node[o] (out) at (15.3,-7.8) {Curated set of $N$ images};
\coordinate (a1) at (4.4,8.1);\coordinate (a2) at (12.7,2.3);
\coordinate (b1) at (4.4,1.6);\coordinate (b2) at (12.7,-3.45);
\coordinate (c1) at (4.4,-4.2);\coordinate (c2) at (12.7,-6.0);
\coordinate (d1) at (4.4,-6.8);\coordinate (d2) at (17.0,-8.8);
\coordinate (i1) at (-1.8,7.55);
\begin{scope}[on background layer]
\node[pan, fit=(tour)(pool)(real)(bud)(i1), inner sep=.3cm] (p0) {};
\node[pan, fit=(u)(wi)(wd)(w)(a1)(a2), inner sep=.2cm] (p1) {};
\node[pan, fit=(si)(sg)(sc)(b1)(b2), inner sep=.2cm] (p2) {};
\node[pan, fit=(fl)(c1)(c2), inner sep=.2cm] (p3) {};
\node[pan, fit=(dr)(out)(d1)(d2), inner sep=.2cm] (p4) {};
\node[draw=acc, dashed, line width=.7pt, rounded corners=6pt, fit=(p1)(p2)(p3)(p4), inner xsep=.25cm, inner ysep=.65cm] (band) {};
\end{scope}
\node[lab, anchor=north west] at ($(p0.north west)+(.15,-.1)$) {Inputs};
\node[lab, anchor=north west] at ($(p1.north west)+(.15,-.1)$) {1  Weights};
\node[lab, anchor=north west] at ($(p2.north west)+(.15,-.1)$) {2  Image score};
\node[lab, anchor=north west] at ($(p3.north west)+(.15,-.1)$) {3  Safety filter};
\node[lab, anchor=north west] at ($(p4.north west)+(.15,-.1)$) {4  Draw};
\node[font=\footnotesize\itshape, color=acc, anchor=south east] at (band.north east) {computed separately for each class};
\draw[a] (tour) -- (u);
\draw[a] (u.east) -- ++(.45,0) |- (wi.west);
\draw[a] (u.east) -- ++(.45,0) |- (wd.west);
\draw[a] (wi.east) -- ++(.6,0) |- (w.east);
\draw[a] (wd.south) -- (w.north);
\draw[a] (pool.east) -- (si.west);
\draw[a] (pool.east) -- ++(.4,0) |- (sg.west);
\draw[a] (real.east) -- (sg.west);
\draw[a] (si.east) -- ++(.5,0) |- (sc.west);
\draw[a] (sg.east) -- ++(.5,0) |- (sc.west);
\draw[a] (w.south) -- (sc.north);
\draw[a] (sc.south) -- (fl.north);
\draw[a] (real.east) -- ++(.8,0) |- (fl.west);
\draw[a] (fl.south) -- (dr.north);
\draw[a] (bud.east) -- (dr.west);
\draw[a] (dr.east) -- (out.west);
\end{tikzpicture}}
\caption{The evaluation algorithm. Per-class weights derived from the tournament combine the instance scores
and the group scores of the pool into one score per image. A safety filter removes images close to the real
training images, and a temperature-scaled draw selects the quota of each class.}
\label{fig:algorithm}
\end{figure}

The tournament of Section~\ref{sec:tournament} shows that the value of a metric depends on the
class. The algorithm turns those measurements into a sampling distribution over the synthetic
pool. It takes four inputs: a pool with per-image metric scores, a weight for every metric in every
class, a budget $N$, and the real training images. The weights are derived from the tournament, from each metric's $F_2$ on the
class (Section~\ref{sec:algorithm-per-class-utility-weights}). The tournament evaluated 16 metrics, and the algorithm uses 15 of them. FID-DINOv2 is left
out because it is the same statistic as FID-Inception computed on a different feature backbone, so
keeping both would give one signal two weights. FID-Inception stays because it is the original and
most widely used FID variant \citep{heusel2017} and scored higher in the tournament
(Section~\ref{sec:tournament}), which leaves three distributional signals: FID, KID and SWD. For each
class $c$ the algorithm then (i) scores every image by combining the weighted per-image scores
with a per-generation-group score for the distributional metrics
(Section~\ref{sec:algorithm-group-scored-distributional-te}), (ii) removes images that lie too
close to a real training image (Section~\ref{sec:algorithm-calibrated-safety-threshold}), and
(iii) draws the class quota from a temperature-scaled softmax over the combined score
(Section~\ref{sec:algorithm-temperature-scaled-sampling-an}). Only the weights come from the
tournament, and the other steps use the pool and the real training images. Every quantity is computed separately for each of the seven classes, and
no class is named in the implementation. Figure~\ref{fig:algorithm} shows the steps.

\subsubsection{Per-Class Utility Weights}
\label{sec:algorithm-per-class-utility-weights}

Let $M$ be the 15-metric portfolio, made up of the 12
instance metrics $M_{\text{inst}}$ and the three distributional metrics $D$ (FID-Inception, KID-DINOv2
and SWD-DINOv2), and let $\text{F2}[c,m]$ be the mean tournament $F_2$ of metric $m$ on class $c$, recomputed from the
saved predictions. Weights are built differently for the two types of metric. Both start from the same
step. Every metric in $M$ receives a utility $u_A$ equal to its distance above the weakest metric on
that class, so no metric receives exactly zero:
\begin{equation}
u_A[c,m] = \bigl(\text{F2}[c,m] - \min_{m' \in M}\text{F2}[c,m']\bigr) + \epsilon,
\label{eq:utility-a}
\end{equation}
with $\epsilon = 10^{-4}$. The utilities are normalized into weights $w_A$:
\begin{equation}
w_A[c,m] = \frac{u_A[c,m]}{\sum_{m' \in M} u_A[c,m']}.
\label{eq:weight-a}
\end{equation}
For an instance metric, $w_A$ is the final weight. The 12 instance metrics combine into a
continuous per-image score with roughly 275,000 distinct values. Therefore, the contribution of a weak
component is reduced relative to the other eleven, and a class can benefit from many of them at
moderate weight. A random anchor would
remove that breadth by concentrating the weight on the few metrics that beat random sampling,
so the instance metrics keep the worst-metric anchor,
which only shrinks a weak metric's weight.

For the three distributional metrics, $w_A$ determines only their combined weight in the class, and a
second step then divides that total among them again. The total is taken from the first step because
the two utilities are on different scales: the clipped utility of Equation~\ref{eq:utility-b} is much
smaller than the utility of Equation~\ref{eq:utility-a}, so normalizing the two families together
would shrink the combined weight of the distributional metrics. This combined weight is the
distributional share of class $c$:
\begin{equation}
\sigma_c = \sum_{d \in D} w_A[c,d].
\label{eq:share}
\end{equation}
The second step keeps $\sigma_c$ and divides it in proportion to how far each metric beats the random
control. The distributional metrics are anchored to random because they act as a vote over 16 discrete
groups in which 17,217 images share a value, so a weak metric carries a third of the pull and moves
thousands of images toward a group that is worse for that class. A metric that does not beat random
sampling on class $c$ therefore receives zero:
\begin{equation}
u_B[c,d] = \max\bigl(0,\ \text{F2}[c,d] - \text{F2}[c,\text{Random}]\bigr).
\label{eq:utility-b}
\end{equation}
These utilities are normalized within $D$ and scaled to the share $\sigma_c$:
\begin{equation}
w_B[c,d] = \sigma_c \, \frac{u_B[c,d]}{\sum_{d' \in D} u_B[c,d']}.
\label{eq:weight-b}
\end{equation}
The final weight $w[c,m]$ is $w_A[c,m]$ for an instance metric and $w_B[c,m]$ for a distributional metric.
The weights of each class sum to one. If no distributional metric beats random on a class, 
the denominator of Equation~\ref{eq:weight-b} is zero and that class keeps its $w_A$ weights. The 
random baseline in Equation~\ref{eq:utility-b} is the 60-seed mean from the tournament. The temperature 
sweep of Section~\ref{sec:results-temperature} used an earlier weighting in which the worst anchor was 
applied to all 15 metrics.

\subsubsection{Combined Image Score}
\label{sec:algorithm-group-scored-distributional-te}

\paragraph{Instance score.}
For an instance metric $m$, the score $S_m(x) \in [0,1]$ of image $x$ is its raw value clipped at the
1st and 99th percentiles of that metric within the class of $x$ and then min-max scaled to $[0,1]$,
inverted when lower is better. Clipping stops one extreme value from compressing the scores of the
others, and min-max scaling gives every score the same range, so that instance scores and group scores
can be combined with the weights. Percentiles are taken within the class because sampling is a softmax
over one class at a time, so only within-class spread determines how sharply the draw discriminates.

\paragraph{Group score.}
FID, KID and SWD describe a set of images, so they cannot score a single image. The algorithm
therefore needs a partition of the pool into groups. It scores each group, and every image takes the
score of its group. The partition is a parameter of the algorithm. In this experiment, a
group is one of the 16 checkpoint-by-guidance configurations, which contains 17,217 images pooled
over all seven classes. Grouping by class, which would match the rest of the algorithm, is not
possible with these data. A class-wise FID would compare the synthetic images of a class with the
real images of that class, and each side needs roughly 2,048 images for a full-rank
$2048 \times 2048$ covariance. However, ISIC-2018 has that many only for NV
(5,364 real training images), and the other six classes have between 92 and 890. This same grouping 
is applied to all three distributional metrics where pooled groups are large enough for all three. 
For group $k$ and distributional metric $d$, let $r_d(k)$ be the raw FID, KID or SWD of the group's full
image set against the real training set. It is computed once and cached, because it depends only on the
pool, the real set and the grouping. The 16 raw values of each metric $d$ are clipped and min-max scaled in the same way, over the 16
groups, and inverted because lower is better, so that the best group receives 1 and the worst 0. This
gives the group score $q_d(k) \in [0,1]$.

\paragraph{Combined score.}
Every image $x$ of class $c$ from group $k(x)$ then receives a combined score, using the final weights $w$ of
Section~\ref{sec:algorithm-per-class-utility-weights}:
\begin{equation}
S_c(x) = \sum_{m \in M_{\text{inst}}} w[c,m]\, S_m(x) \;+\; \sum_{d \in D} w[c,d]\, q_d\bigl(k(x)\bigr).
\label{eq:composite}
\end{equation}
Every instance score $S_m(x)$ and group score $q_d(k)$ lies in $[0,1]$, and the weights of a class sum
to one across all 15 metrics, so $S_c(x)$ is a weighted average of values in $[0,1]$ and lies in
$[0,1]$. The temperature of Section~\ref{sec:algorithm-temperature-scaled-sampling-an} therefore has
the same meaning for every class.
A class whose weights lean toward the distributional metrics has its draw pulled toward the best
generation groups, and a class that leans toward instance metrics does not. Which classes
concentrate on which groups follows from the weights alone.

\subsubsection{Safety Filter}
\label{sec:algorithm-calibrated-safety-threshold}

\paragraph{Filter.}
Before sampling, the algorithm removes every synthetic image whose DINOv2 cosine similarity to its
nearest real training image is at least $\tau$, so a higher $\tau$ removes fewer images. The threshold
is the same for every class, and a per-class quota guard prevents the filter from reducing a class
below the count its draw needs.

\paragraph{Calibration.}
Choosing $\tau$ requires knowing what a given cosine value means, and no labeled copies of training
images exist, so $\tau$ is calibrated on real images. ISIC photographed many lesions more than once, and
images of one lesion share a \texttt{lesion\_id}, so two real training images with the same identifier
show the same physical lesion. This is a known-positive example of the relation a memorized generation
has to its source. We enumerated all pairs of real training images with cosine similarity of at least
0.90. Pairs below 0.90 were not enumerated, because they are very numerous and rarely show the same
lesion. This gives 117,751 candidate pairs, of which 776 (0.66\%) share a lesion identifier and form the set $S$
of same-lesion pairs. For a threshold $\tau$, let $C(\tau)$ be the set of candidate pairs with cosine
similarity at least $\tau$. We compute three quantities for each candidate threshold. The first is the
precision, the share of the candidate pairs at or above the threshold that are same-lesion pairs:
\begin{equation}
\text{precision}(\tau) = \frac{|C(\tau) \cap S|}{|C(\tau)|}.
\label{eq:precision}
\end{equation}
The second is the recall, the share of all same-lesion pairs that reach the threshold:
\begin{equation}
\text{recall}(\tau) = \frac{|C(\tau) \cap S|}{|S|}.
\label{eq:recall}
\end{equation}
A low precision means that many flagged pairs are not copies, and a low recall means that many
same-lesion pairs are missed. The third quantity is the number of synthetic pool images that the filter
would remove at that threshold. Plotted against the threshold, these three quantities form the
calibration curve.

Four thresholds are evaluated throughout the paper, $\tau \in \{0.90, 0.93, 0.97, 0.99\}$, each chosen
from its own property of the data rather than a single shared rule. The value 0.90 is where every
same-lesion pair is flagged (recall 100\%, by construction of the enumeration lower bound) and sits close
to the real training set's own nearest-neighbor similarity (mean cosine 0.9047,
Section~\ref{sec:safety-feature-space-memorization-sco}); precision there is only 0.7\%, so below
this point the filter can no longer distinguish memorization from ordinary inter-patient
resemblance. The value 0.93 is the highest tested value with recall above 50\% (55.0\%; the next
step, 0.94, drops to 41.5\%), the midpoint of the detector's tested sensitivity range. The value
0.97 is the lowest tested value with precision above 95\% (96.2\% specifically), the conventional
high-confidence cutoff. The value 0.99 is the point at which the filter removes none of the
275,472 pool images, an unfiltered reference point rather than a calibrated cut. The
resulting curve is in Section~\ref{sec:results-calibration}.

\subsubsection{Sampling and Budget}
\label{sec:algorithm-temperature-scaled-sampling-an}

The budget $N$ is a required input, and it is divided among the classes by target fractions $\rho_c$.
The quota of class $c$ is $n_c = \lfloor \rho_c N \rfloor$, with any remainder added to the largest
class. All experiments set $\rho_c$ to the class proportions of the real training set. The algorithm
draws $N$ images from the scored pool for a specified sampling budget $N$, and the benchmark of
Section~\ref{sec:benchmark} reports $N = 8{,}012$, which matches the real training set size so that
cohorts are compared at equal volume, and $N = 20{,}000$.

Each class $c$ draws $n_c$ images without replacement from its set of candidates $C_c$, the pool
images of class $c$ that remain after the safety filter of Section~\ref{sec:algorithm-calibrated-safety-threshold}.
Each candidate $x$ is drawn with probability
\begin{equation}
p(x) \;=\; \frac{\exp\!\bigl(S_c(x)/T\bigr)}{\sum_{x' \in C_c} \exp\!\bigl(S_c(x')/T\bigr)}.
\label{eq:softmax}
\end{equation}
The exponential makes the temperature meaningful: with weights proportional to $S_c(x)$ itself, $T$
would cancel in the normalization. With it, the relative odds of two images depend only on the gap
$\Delta$ between their scores and equal $\exp(\Delta/T)$, so $T$ sets how strongly a given gap favors
the higher-scoring image. Low $T$ concentrates the draw on the highest-scoring images, and high $T$
approaches uniform sampling. The locked temperature is $T=0.05$ (Section~\ref{sec:results-temperature}).

\subsection{Feature-Space Memorization Score}
\label{sec:methods-fsms}

Curation that favors images close to the real data could, as a side effect, favor images that
reproduce individual training photographs. Diffusion models are known to emit near-copies of their
training images \citep{carlini2023, akbar2023, dar2025}, which is a privacy risk for patient
images. This subsection defines the safety measure used throughout the paper. Section~\ref{sec:safety-feature-space-memorization-sco}
tests that the measure behaves as intended, and Section~\ref{sec:safety-threshold-effect} reports how safety
and utility change as the safety threshold $\tau$ of Section~\ref{sec:algorithm-calibrated-safety-threshold}
is tightened.

\paragraph{Definition.}
Let $f(x)$ be the DINOv2 (base) CLS embedding of an image standardized to $512 \times 512$
pixels, let $R$ be the 8{,}012 real training images, and let $S$ be a set of synthetic images. The
Feature-Space Memorization Score (FSMS) is
\begin{equation}
\text{FSMS}(S,R) \;=\; 1 \;-\; \frac{1}{|S|}\sum_{x \in S}\; \max_{y \in R}\; \cos\bigl(f(x), f(y)\bigr).
\label{eq:fsms}
\end{equation}
Subtracting the mean similarity from 1 turns it into a distance. A low value means that synthetic
images lie close to specific training images, and a high value means they lie far from every training
image, so FSMS is a safety score in which higher is safer. It is computed over the synthetic images only. The per-image term
$s(x)=\max_{y\in R}\cos(f(x),f(y))$ is also what the safety filter of
Section~\ref{sec:algorithm-calibrated-safety-threshold} thresholds. FSMS is a self-defined statistic
without a natural scale, so the checks below supply reference points.

\subsection{Benchmark Protocol}
\label{sec:benchmark-protocol}

The tournament and the algorithm were developed on a validation split, so their scores are
selection results. The confirmatory test is described here and its results are in Section~\ref{sec:benchmark}. Curated synthetic images are added
to the real training set, classifiers are trained, and every classifier is scored on the official
ISIC-2018 Task 3 test set, which no earlier step in this paper used.

\subsubsection{Experimental Design}
\label{sec:benchmark-experimental-design}

\paragraph{Task.}
Each classifier is trained on the 8,012 real training images together with a curated synthetic
set. The classifier, the frozen ImageNet backbone with a trained linear head, and all
optimization settings are those of Section~\ref{sec:tournament-tournament-protocol}. The
validation split of 2,003 real images selects the early-stopping epoch. Models are scored on the
1,512-image test set (AKIEC 43, BCC 93, BKL 217, DF 44, MEL 171, NV 909, VASC 35 images).
The test set was never used to build the tournament weights, to select the algorithm
configuration, for early stopping, or for any other choice.

\paragraph{Locked model variance.}
Weight initialization, data-loader seed and CUDA convolution algorithms are fixed in every run
(seed 42). Within a cohort, the only quantity that changes between runs is which synthetic images
the curation procedure selected, using 30 independent curation seeds. The spread of a cohort
therefore measures sensitivity to the selected data. It does not include classifier initialization
variance. The fixed seed does not make runs bit-reproducible: PyTorch's full deterministic mode was
not enabled, so some graphics processing unit (GPU) matrix operations can differ between executions of the same
configuration (Section~\ref{sec:limitations}).

\paragraph{Statistics.}
Cohorts use different curation procedures, so runs are not paired and all comparisons are
unpaired: Welch's $t$-test, with the Mann-Whitney $U$ test (MWU) as a distribution-free check and
Cohen's $d$ with a pooled standard deviation as effect size. Differences are reported with 95\% confidence intervals (CI). The primary comparisons are the
algorithm against the FID-Inception best-metric cohort at $N=8{,}012$, on two outcomes: test
Macro-$F_2$ (utility) and FSMS (safety). Each is run once for each
of four safety thresholds (Section~\ref{sec:algorithm-calibrated-safety-threshold}), and
Bonferroni correction is applied across the four tests of each outcome. All other comparisons are secondary and
uncorrected. The real-data-only classifier has no sampling step and is trained once, so it has no
spread and cannot be tested.

\subsubsection{Cohorts}
\label{sec:benchmark-cohorts}

Every synthetic cohort adds its images to the same 8,012 real images.
\begin{itemize}
\item \textbf{Real data only} is trained once and has no synthetic images.
\item \textbf{Random} draws a class-stratified sample uniformly from the full 275,472-image pool.
\item \textbf{Best-metric (FID-Inception)} represents the common practice of selecting the
generation configuration with the best FID-Inception against the real data and sampling from it.
The ranked-best configuration is the 25k-kimg checkpoint at CFG scale 1.0. Samples are
class-stratified and uniformly random within that configuration.
\item \textbf{Algorithm} is the locked configuration of Section~\ref{sec:algorithm}: weights
derived from the tournament, group-scored distributional term,
$T=0.05$, and a safety threshold $\tau$. Results are reported at four thresholds, $\tau \in
\{0.90, 0.93, 0.97, 0.99\}$.
\end{itemize}
All synthetic cohorts follow the real class proportions. Cohorts of $N=8{,}012$ match the real
training set one to one. Each synthetic cohort has 30 runs, so the Swin-B benchmark comprises 241
classifiers: one real-only run and eight synthetic cohorts of 30 (random sampling, the best-metric
cohort and the algorithm at four thresholds at $N=8{,}012$, and random sampling and the algorithm at
$N=20{,}000$).

\paragraph{Larger volume.}
A second volume, $N=20{,}000$, tests the dose response: whether the gains of the algorithm hold when it
can curate a larger set. Here 20{,}000 synthetic images are added to the 8{,}012 real images, giving
28{,}012 training images, and the cohorts are Random and the algorithm only. The best-metric cohort has no
counterpart at this volume. It draws every image from one generation configuration, which holds 6{,}501 NV
images. That covers the 5{,}364 NV images that $N=8{,}012$ requires and falls short of the roughly 13{,}390
that $N=20{,}000$ requires. The algorithm was run at one threshold, $\tau=0.97$, the lowest tested value
with precision above 95\%.

\section{Results}
\label{sec:results}

\subsection{Tournament Ranking}
\label{sec:tournament-per-class-results}
\label{sec:tournament}

\begin{figure}[tbp]
\centering
\includegraphics[width=\linewidth]{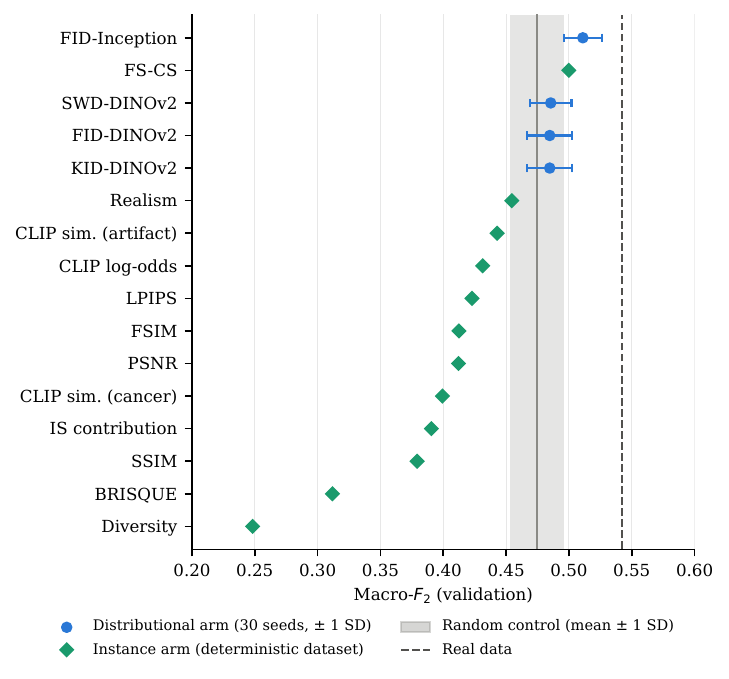}
\caption{Macro-$F_2$ of every tournament arm on the validation split, sorted by the Swin-B score.
Distributional arms and the random control are means over seeds (the bar is one standard deviation),
and instance arms are single deterministic datasets. The exact values are in Appendix~\ref{app:tournament}.}
\label{fig:tournament-ranking}
\end{figure}

Figure~\ref{fig:tournament-ranking} ranks every arm by Macro-$F_2$. On Swin-B, FID-Inception
is the strongest synthetic arm at 0.511, against 0.475 for random sampling and 0.542 for real data
only. It beats random sampling by 0.036, the largest gain of any distributional arm. The DINOv2 arms
gain about 0.01, a much smaller difference (Table~\ref{tab:tournament-paired}). FID-DINOv2 and
KID-DINOv2 select the same configuration
and are sampled with the same seeds, so their curated sets and results are identical. Among instance metrics, FS-CS is the only arm above
random (0.500), and the remaining eleven fall below it. Each instance arm is one deterministic dataset with one score, so it has no seed-to-seed spread of its own, and gaps
of 0.01 to 0.02 to random fall inside the spread of random draws (standard deviation 0.021), so no
instance comparison is interpretable on its own.

\begin{table}[t]
\centering
\caption{Difference in Swin-B Macro-$F_2$ between each distributional arm (30 seeds) and the random
control (60 runs), with the $p$-value of Welch's $t$-test.}
\label{tab:tournament-paired}
\small
\begin{tabular}{@{}lrr@{}}
\toprule
Arm & $\Delta$ & $p$ \\
\midrule
FID-Inception & +0.036 & $4.4\times10^{-14}$ \\
SWD-DINOv2 & +0.011 & 0.011 \\
FID-DINOv2 & +0.010 & 0.024 \\
KID-DINOv2 & +0.010 & 0.024 \\
\bottomrule

\end{tabular}
\end{table}

\subsection{No Single Metric Serves Every Class}
\label{sec:tournament-no-single-metric-serves-every-class}

\begin{figure}[tbp]
\centering
\includegraphics[width=\linewidth]{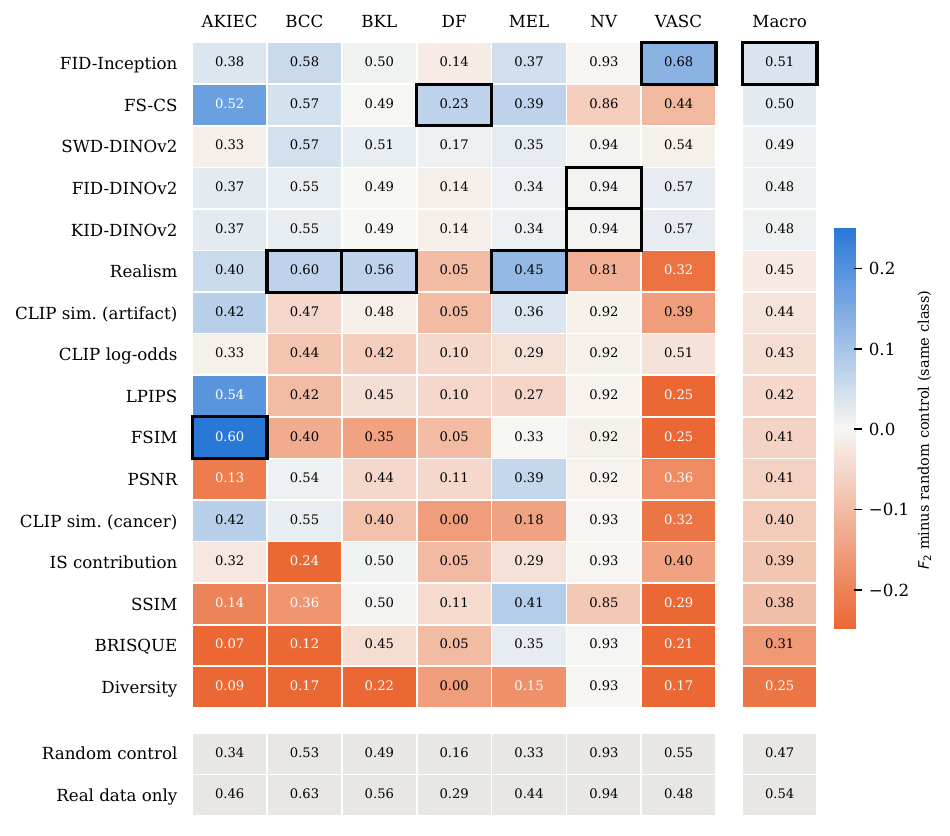}
\caption{Per-class $F_2$ of every arm on the validation split for Swin-B. Cell color is the
difference from the random control in the same class, and the number is the arm's $F_2$. A black
frame marks the best arm in each column. Distributional arms are 30-seed means and instance arms
are single deterministic datasets. The exact values are in Appendix~\ref{app:tournament}.}
\label{fig:tournament-heatmap}
\end{figure}

\begin{figure}[tbp]
\centering
\includegraphics[width=\linewidth]{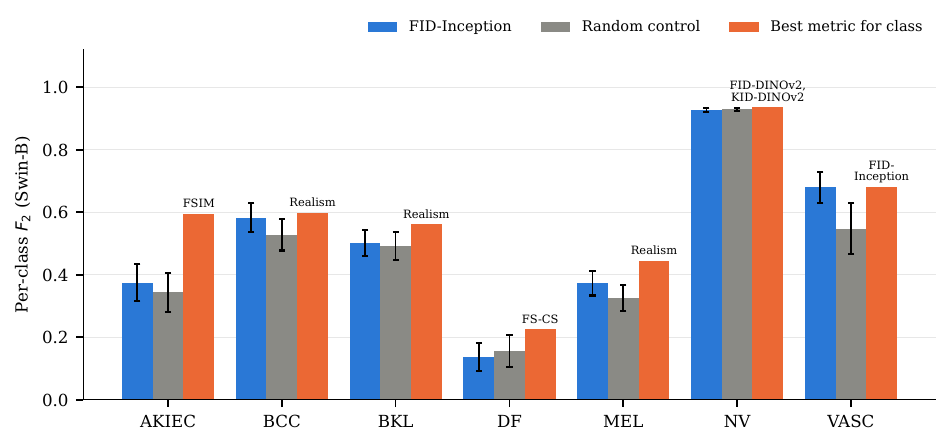}
\caption{Per-class $F_2$ on the validation split for Swin-B. Error bars are one standard
deviation over seeds (30 for FID-Inception, 60 for the random control). The best-metric bar is
a single deterministic dataset except on NV, where the winning DINOv2 arm is a 30-seed mean.}
\label{fig:tournament-per-class}
\end{figure}

Figure~\ref{fig:tournament-heatmap} breaks the Swin-B scores down by class, recomputed from each
run's saved predictions. The best arm differs from class to class (Figure~\ref{fig:tournament-per-class}). FSIM leads on AKIEC (0.596), Realism on BCC (0.597),
BKL (0.561) and MEL (0.445), FS-CS on DF (0.225), FID-Inception on VASC (0.680), and FID-DINOv2 and
KID-DINOv2 on NV (0.936). Five different selection rules therefore lead across seven classes.

FID-Inception, the usual choice, reaches 0.680 on VASC and falls to 0.137 on DF. On AKIEC it scores
0.375 where FSIM scores 0.596. Against random sampling
(Table~\ref{tab:tournament-fid-per-class}), FID-Inception is clearly better on VASC and also better
on BCC and MEL, and the gain on AKIEC is small. It shows no meaningful difference on BKL or NV (on NV
the difference is 0.003). On DF it scores 0.019 below the random mean, a gap that does not reach
significance, so the supported statement is that FID-Inception offers no measurable advantage on DF.

\begin{table}[t]
\centering
\caption{Swin-B per-class $F_2$ of FID-Inception (30 seeds) and of the random control (60 runs), their
difference, and the $p$-value of Welch's $t$-test.}
\label{tab:tournament-fid-per-class}
\small
\begin{tabular}{@{}lrrrr@{}}
\toprule
Class & FID-Inception & Random & $\Delta$ & $p$ \\
\midrule
AKIEC & 0.375 & 0.343 & +0.032 & 0.021 \\
BCC & 0.583 & 0.528 & +0.055 & $2.1\times10^{-6}$ \\
BKL & 0.502 & 0.492 & +0.009 & 0.319 \\
DF & 0.137 & 0.156 & $-$0.019 & 0.080 \\
MEL & 0.373 & 0.326 & +0.048 & $1.7\times10^{-6}$ \\
NV & 0.927 & 0.930 & $-$0.003 & 0.013 \\
VASC & 0.680 & 0.548 & +0.132 & $4.6\times10^{-15}$ \\
\midrule
Macro & 0.511 & 0.475 & +0.036 & $4.4\times10^{-14}$ \\
\bottomrule

\end{tabular}
\end{table}

Class size does not account for the pattern. Distributional arms lead on the largest class (NV,
5,364 real training images) and on the second smallest (VASC, 114), while an instance metric
leads on DF (92), which is smaller than both. The evidence behind the cells differs in strength.
Distributional entries rest on 30 seeds, and instance entries are single deterministic datasets, so they are
indicative. The validation split also contains only 23 DF and 28 VASC images, so one image
shifts the recall of those classes by about 4 points. The central observation does not depend on any
single instance cell. FID-Inception is a 30-seed arm, and its scores range from clearly above
random on some classes to indistinguishable from it on others. A single global choice of metric
therefore cannot be best for every class. This motivates estimating curation weights per class
from these measurements, as described in Section~\ref{sec:algorithm}.

\subsection{Feature Backbone of the Metrics}
\label{sec:feature-backbone}

The tournament computes FID on two feature backbones, InceptionV3 and DINOv2, so that the backbone can
be compared with the statistic held fixed. On Swin-B, FID-Inception reaches a Macro-$F_2$ of 0.511,
against 0.485 for FID-DINOv2 (a difference of 0.026 over the 30 seeds, Welch's $p<0.001$;
Table~\ref{tab:backbone}). FID-Inception clearly beats random sampling (0.475), whereas FID-DINOv2 is
only about 0.01 above it, a difference with Welch's $p=0.024$. On ResNet-50 FID-Inception is ahead by
0.008 ($p=0.18$), and both are above random sampling (0.460), so the two backbones cannot be separated on
that architecture. Each arm samples only from the configuration its metric ranks best. FID-Inception and FID-DINOv2 rank different
configurations best (25k kimg at guidance 1.0 and 1.5), so the comparison of backbones is a comparison of those
two configurations. Section~\ref{sec:discussion-feature-spaces} discusses this.

\begin{table}[t]
\centering
\caption{FID on the two feature backbones: the configuration each arm selects and its validation
Macro-$F_2$ on each architecture (mean $\pm$ standard deviation over 30 curation seeds). The random
control has 60 seeds on Swin-B and 30 on ResNet-50.}
\label{tab:backbone}
\small
\begin{tabular}{@{}llllll@{}}
\toprule
Arm & Features & Selected configuration & Swin-B & ResNet-50 \\
\midrule
FID-Inception & InceptionV3 & 25k kimg, guidance 1.0 & $0.511 \pm 0.015$ & $0.500 \pm 0.018$ \\
FID-DINOv2 & DINOv2 & 25k kimg, guidance 1.5 & $0.485 \pm 0.018$ & $0.492 \pm 0.027$ \\
\midrule
Random control & --- & --- & $0.475 \pm 0.021$ & $0.460 \pm 0.032$ \\
\bottomrule

\end{tabular}
\end{table}

\subsection{Evidence for the Design Choices}
\label{sec:design-evidence}

This subsection reports the evidence for the design choices of Section~\ref{sec:algorithm}: the resulting weights, the calibration of the safety threshold and the temperature sweep. These are validation-split selection results, and the confirmatory test is Section~\ref{sec:benchmark}.

\subsubsection{Metric Weights}
\label{sec:algorithm-weights-results}

\begin{figure}[tbp]
\centering
\includegraphics[width=0.82\linewidth]{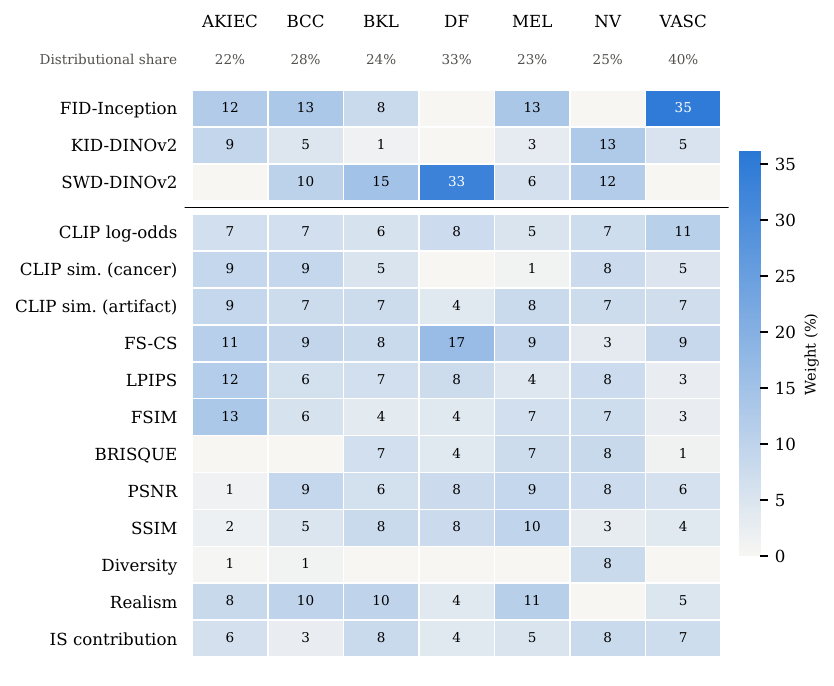}
\caption{Locked Swin-B metric weights $w[c,m]$ per class, in percent. Each column sums to 100, and the
number above it is the distributional share $\sigma_c$, the weight going to the three distributional
metrics. Blank cells have a weight
below 0.5\%. Exact values are in Appendix~\ref{app:weights}.}
\label{fig:weights}
\end{figure}

Figure~\ref{fig:weights} shows the resulting Swin-B weights $w[c,m]$, and
Appendix~\ref{app:weights} lists the exact values. The distributional share ranges from 21.6\% (AKIEC)
to 40.4\% (VASC). The VASC weight is concentrated on FID-Inception, which holds 86.7\% of that
share. DF puts its entire distributional share on SWD-DINOv2 and gives 16.6\% to FS-CS, the
best single metric for that class in the tournament. SWD-DINOv2 holds the largest cell on DF because it is the only
distributional metric above the random control there, so the whole share lands on it. AKIEC spreads weight across 13 metrics, none
above 13.3\%.

Because Equation~\ref{eq:utility-b} is a hard threshold, a distributional metric close to the random
control can be included or excluded by the noise in the baseline estimate. A weight of zero therefore
means that the metric did not clear a noisy threshold.

\subsubsection{Safety Threshold Calibration}
\label{sec:results-calibration}

\begin{figure}[tbp]
\centering
\includegraphics[width=\linewidth]{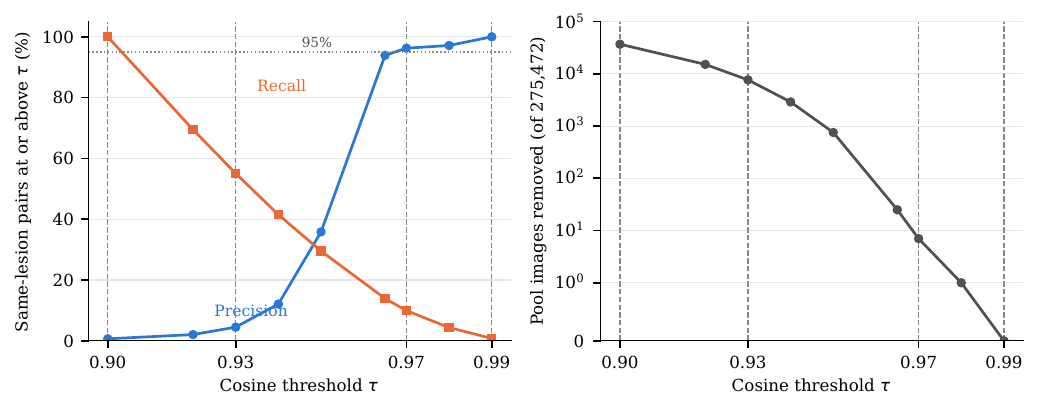}
\caption{Calibration of the safety threshold on same-lesion pairs of real training images. Left:
precision and recall of the same-lesion label among pairs at or above each cosine threshold. Right:
synthetic pool images (of 275{,}472) at or above each threshold, on a symmetric logarithmic axis.
Dashed lines mark the four thresholds used in the paper.}
\label{fig:calibration}
\end{figure}

Figure~\ref{fig:calibration} shows the precision, the recall and the number of pool images removed
at each threshold. Precision stays below 36\% for $\tau \le 0.95$, rises to 93.9\% at 0.965 and 96.2\% at 0.97,
and reaches 100\% at 0.99 (6 of 6 pairs). At $\tau=0.90$ the filter removes 36,929 pool images
(13.4\%). At 0.93, where recall is 55.0\%, it removes 7,666 (2.8\%). At 0.97, where precision is 96.2\%,
it removes 7. At 0.99, no pool image reaches the threshold, so nothing is removed.

Two limits apply. The positives are re-photographs of one lesion, which differ from diffusion
reproductions, so the calibration bounds false removals and says nothing about missed
reproductions. The recall column also shows that a threshold with high precision identifies only
a small share of same-lesion pairs (9.9\% at 0.97).

\subsubsection{Temperature}
\label{sec:results-temperature}

The temperature was set by a sweep over three values with the worst anchor applied to all 15 metrics
(Table~\ref{tab:temperature}). The mean rises from $T=0.1$ to $T=0.05$ and falls at $T=0.02$, so the
maximum of the tested range lies at $T=0.05$. Neither neighboring difference is significant, so
the three values perform similarly, and we use $T=0.05$ for its highest mean.

\begin{table}[t]
\centering
\caption{Validation Macro-$F_2$ of the algorithm at three temperatures (Swin-B, worst anchor applied
to all 15 metrics, mean $\pm$ standard deviation over 30 curation seeds).}
\label{tab:temperature}
\small
\begin{tabular}{@{}lc@{}}
\toprule
$T$ & Macro-$F_2$ \\
\midrule
0.1 & $0.5174 \pm 0.0148$ \\
0.05 & $0.5218 \pm 0.0187$ \\
0.02 & $0.5150 \pm 0.0161$ \\
\bottomrule
\end{tabular}
\end{table}

\subsection{Clinical Augmentation Benchmark}
\label{sec:benchmark}

\subsubsection{Algorithm Against the Baselines}
\label{sec:benchmark-against-baselines}

\begin{figure}[tbp]
\centering
\includegraphics[width=\linewidth]{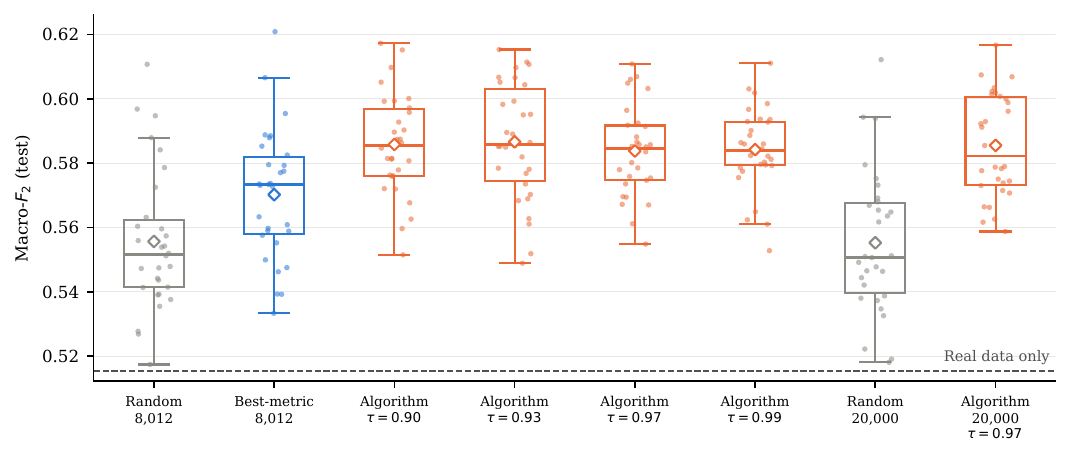}
\caption{Macro-$F_2$ on the test set for Swin-B, one point per curation seed (30 per cohort).
Boxes show the interquartile range and median, and diamonds mark the mean. The four algorithm
cohorts at $N=8{,}012$ are the same algorithm at four safety thresholds $\tau$. The $N=20{,}000$
algorithm cohort was run at $\tau=0.97$ only. The dashed line is the single real-data-only run.}
\label{fig:benchmark-swin}
\end{figure}

Table~\ref{tab:cohorts-swin} and Figure~\ref{fig:benchmark-swin} give the Swin-B results. All
synthetic cohorts improve on the single real-only run (Macro-$F_2$ 0.5154), and every one of the
30 runs in every synthetic cohort lies above it. Because the real-only run has no replicates,
this comparison is descriptive.

\begin{table}[t]
\centering
\caption{Swin-B test-set results for real-plus-synthetic training. Entries are mean $\pm$
standard deviation over 30 curation seeds. $\tau$ is the safety threshold. FSMS measures distance from real training images
(higher is safer, Section~\ref{sec:methods-fsms}) and is not defined without synthetic images.}
\label{tab:cohorts-swin}
\small
\begin{tabular}{@{}lrrcc@{}}
\toprule
Cohort & Synthetic $N$ & Runs & Macro-$F_2$ & FSMS \\
\midrule
Real data only & --- & 1 & 0.5154 & --- \\
\midrule
Random & $8{,}012$ & 30 & 0.5557 $\pm$ 0.0222 & 0.1639 $\pm$ 0.0009 \\
Best-metric (FID-Inception) & $8{,}012$ & 30 & 0.5702 $\pm$ 0.0202 & 0.1157 $\pm$ 0.0003 \\
\midrule
Algorithm, $\tau=0.90$ & $8{,}012$ & 30 & 0.5858 $\pm$ 0.0158 & 0.1443 $\pm$ 0.0004 \\
Algorithm, $\tau=0.93$ & $8{,}012$ & 30 & 0.5866 $\pm$ 0.0183 & 0.1232 $\pm$ 0.0004 \\
Algorithm, $\tau=0.97$ & $8{,}012$ & 30 & 0.5838 $\pm$ 0.0140 & 0.1152 $\pm$ 0.0004 \\
Algorithm, $\tau=0.99$ & $8{,}012$ & 30 & 0.5842 $\pm$ 0.0128 & 0.1153 $\pm$ 0.0003 \\
\midrule
Random & $20{,}000$ & 30 & 0.5553 $\pm$ 0.0224 & 0.1638 $\pm$ 0.0006 \\
Algorithm, $\tau=0.97$ & $20{,}000$ & 30 & 0.5855 $\pm$ 0.0162 & 0.1174 $\pm$ 0.0002 \\
\bottomrule

\end{tabular}
\end{table}

\begin{figure}[tbp]
\centering
\includegraphics[width=\linewidth]{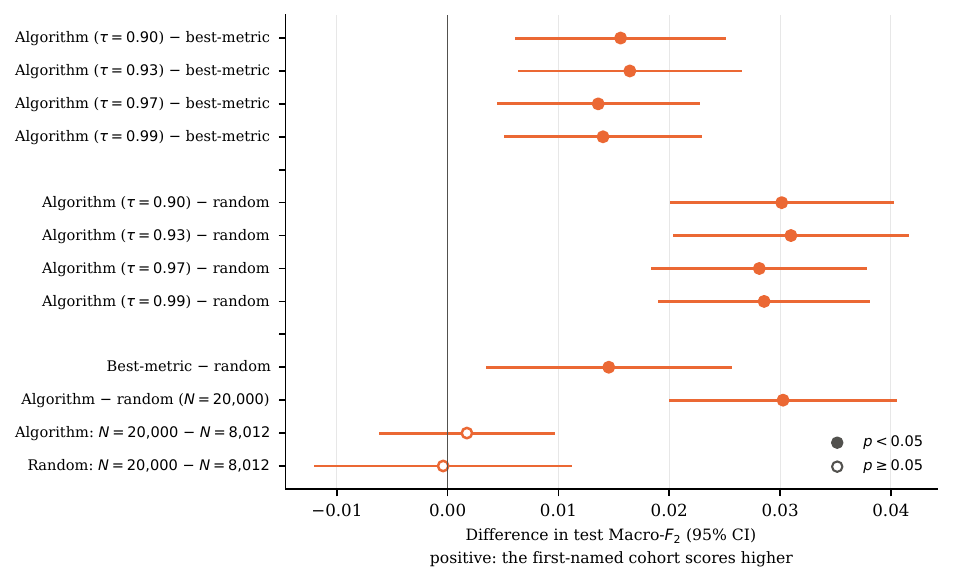}
\caption{Differences in test Macro-$F_2$ between cohorts (the first-named cohort minus the second), with
95\% confidence intervals from Welch's $t$-test. A filled marker means $p<0.05$ and a hollow
marker means $p\geq0.05$. The
algorithm is shown at each of the four safety thresholds.}
\label{fig:forest}
\end{figure}

\paragraph{Algorithm against FID-Inception.}
At each of the four thresholds the algorithm exceeds the best-metric cohort (0.5702) by between
$+0.0136$ and $+0.0164$ Macro-$F_2$. The cohort means are 0.5858 ($\tau=0.90$), 0.5866 (0.93), 0.5838 (0.97) and
0.5842 (0.99). Every comparison is significant with a large effect size and survives Bonferroni correction
over the four tests (Table~\ref{tab:primary} and Figure~\ref{fig:forest}), and the Mann-Whitney tests agree
(Appendix~\ref{app:benchmark-tables}). Table~\ref{tab:primary} also gives the FSMS comparison, which
Section~\ref{sec:safety-threshold-effect} discusses. The four
threshold cohorts are four settings of one algorithm and not independent replications: they share
curation seeds, and the filter removes 7 images at 0.97 and none at 0.99, so those two cohorts
differ only slightly. The algorithm cohorts are also less variable across seeds than the
best-metric cohort (standard deviations 0.013 to 0.018, against 0.020).

\begin{table}[t]
\centering
\caption{Primary comparisons: Swin-B test Macro-$F_2$ and FSMS of the algorithm minus the best-metric
(FID-Inception) cohort at $N=8{,}012$, 30 curation seeds each, at each safety threshold $\tau$. A positive
difference means higher utility for Macro-$F_2$ and a safer set for FSMS. The 95\% confidence interval
and $t$ are from Welch's $t$-test, the adjusted $p$ is Bonferroni-corrected over the four tests of each
outcome, and $d$ is Cohen's $d$.}
\label{tab:primary}
\small
\setlength{\tabcolsep}{4pt}
\begin{tabular}{@{}llrcrrrr@{}}
\toprule
Outcome & $\tau$ & $\Delta$ & 95\% CI & $t$ & $p$ & Adj.\ $p$ & $d$ \\
\midrule
Macro-$F_2$ & 0.90 & +0.0156 & [+0.0062, +0.0250] & 3.33 & 0.0016 & 0.0062 & 0.86 \\
 & 0.93 & +0.0164 & [+0.0065, +0.0264] & 3.30 & 0.0016 & 0.0066 & 0.85 \\
 & 0.97 & +0.0136 & [+0.0046, +0.0226] & 3.02 & 0.0039 & 0.0155 & 0.78 \\
 & 0.99 & +0.0140 & [+0.0052, +0.0228] & 3.21 & 0.0024 & 0.0095 & 0.83 \\
\midrule
FSMS & 0.90 & +0.0287 & [+0.0285, +0.0289] & 317.08 & $<10^{-6}$ & $<10^{-6}$ & 81.87 \\
 & 0.93 & +0.0075 & [+0.0073, +0.0077] & 84.15 & $<10^{-6}$ & $<10^{-6}$ & 21.73 \\
 & 0.97 & $-$0.0004 & [$-$0.0006, $-$0.0002] & $-$4.59 & $2.8\times10^{-5}$ & $1.1\times10^{-4}$ & $-$1.19 \\
 & 0.99 & $-$0.0003 & [$-$0.0005, $-$0.0002] & $-$3.92 & $2.4\times10^{-4}$ & $9.6\times10^{-4}$ & $-$1.01 \\
\bottomrule

\end{tabular}
\end{table}

\paragraph{Other cohorts.}
Best-metric sampling improves on random sampling by $+0.0145$, so single-metric curation helps. The
algorithm improves on random sampling by $+0.028$ to $+0.031$ at the four thresholds, about twice
that gain (Figure~\ref{fig:forest}). Per-class results are in Appendix~\ref{app:benchmark-tables}.

\subsubsection{Effect of the Safety Threshold}
\label{sec:safety-threshold-effect}

\begin{figure}[tbp]
\centering
\includegraphics[width=\linewidth]{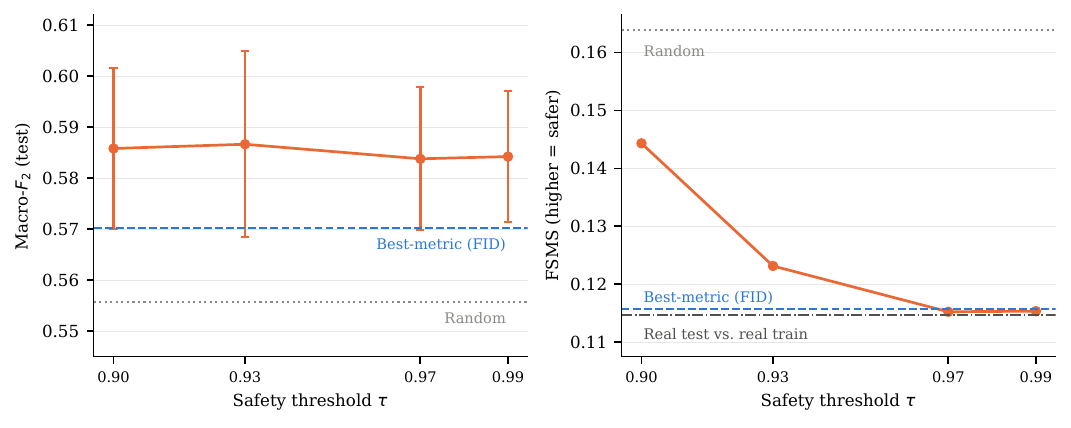}
\caption{Test Macro-$F_2$ (left, mean $\pm$ one standard deviation over 30 seeds) and FSMS (right)
of the algorithm as the safety threshold $\tau$ varies. The dashed line is the best-metric (FID)
cohort, the dotted line is random sampling, and the dash-dot line in the right panel is the FSMS of
held-out real test images against the training set.}
\label{fig:safety-vs-tau}
\end{figure}

The safety threshold $\tau$ removes pool images whose similarity to a training image is at least
$\tau$ before sampling. We ran the algorithm at four values (0.90,
0.93, 0.97 and 0.99), each for Swin-B at $N=8{,}012$ with 30 curation seeds. Each threshold was
chosen from a property of the calibration curve (Section~\ref{sec:results-calibration}).
Table~\ref{tab:cohorts-swin} gives the values and Table~\ref{tab:primary} the tests against the FID
cohort for both outcomes. Figure~\ref{fig:safety-vs-tau} shows the two outcomes against $\tau$:
Macro-$F_2$ stays flat across the four values while FSMS rises as the threshold tightens.

\paragraph{Utility.}
Macro-$F_2$ does not change with $\tau$ in this range. The four means lie between 0.5838 and 0.5866,
within one seed-to-seed standard deviation (0.013 to 0.018), so the data rule out a large utility cost
of tightening $\tau$.

\paragraph{Safety.}
FSMS is flat between 0.99 (0.1153) and 0.97 (0.1152) and rises as $\tau$ tightens further, to 0.1443
at 0.90, as the filter
removes more pool images (7 at 0.97, 36{,}929 or 13.4\% at 0.90). It crosses the FID cohort's
value (0.1157) between 0.93 and 0.97. The algorithm is safer than the FID cohort at 0.90 and 0.93
($+0.0287$ and $+0.0075$) and slightly less safe at 0.97 and 0.99 (by 0.0004 and 0.0003), and all four
differences are significant after Bonferroni correction (Table~\ref{tab:primary}). The shortfall is small. In the detection experiment, planting 1\% of exact copies lowers FSMS by 0.0013,
and the drop is proportional to the planted fraction, so a deficit of 0.0004 has the effect of replacing
about 0.3\% of the images, roughly 25 of 8{,}012, with exact copies. It is about one seed-to-seed
standard deviation of a cohort. At no tested threshold does the algorithm reach random
sampling (0.1639); the closest, $\tau=0.90$, is 0.0195 below it, and at 0.97 it is 0.0486 below.
For reference, held-out real images score 0.1147 against the training set
(Table~\ref{tab:fsms-reference}), so the FID cohort
(0.1157) and the algorithm at 0.97 and 0.99 (0.1152, 0.1153) sit close to the level of real
images the generator never saw.

\paragraph{Curated sets score below random sampling.}
For the FID cohort the cause of the lower FSMS is
visible: it draws every image from the configuration with the lowest FSMS in the pool
(Table~\ref{tab:fsms-pool}). For the algorithm we have not isolated the cause. Curation toward images
that resemble the real data is consistent with the result, since the images closest to the real data
are the ones FSMS penalizes. The filter removes only 7 pool images at 0.97 and none at 0.99, so at
those two thresholds the algorithm is close to unfiltered curation. A tighter $\tau$ raises FSMS with
no measured change in Macro-$F_2$, so utility and safety do not trade off over the tested range.

\subsubsection{Dose-Response at $N=20{,}000$}
\label{sec:benchmark-dose-response-at-n20000}

At $N=20{,}000$ (Section~\ref{sec:benchmark-cohorts}) and $\tau=0.97$ the algorithm scores 0.5855 against 0.5553 for random sampling, a gain
of $+0.0303$ that is close to the $+0.0281$ at $N=8{,}012$ (Figure~\ref{fig:forest}). Neither
cohort changes significantly with $N$: the algorithm gains $0.0017$ and random sampling loses
$0.0004$. The gain over random sampling therefore holds at the larger $N$, and neither cohort at 
$N=20{,}000$ benefits significantly from the extra synthetic images. The algorithm at $N=20{,}000$ also exceeds the
best-metric cohort at $N=8{,}012$ by $0.0153$, although the two differ in $N$. Its FSMS at
$N=20{,}000$ is 0.1174, compared with 0.1152 at $N=8{,}012$ ($\tau=0.97$) and 0.1638 for random
sampling.

\subsection{Validation of the Memorization Score}
\label{sec:safety-feature-space-memorization-sco}

\paragraph{Reference points.}
Scoring the real training set against itself gives exactly 0 by construction, which verifies the
implementation (Table~\ref{tab:fsms-reference}). Scoring each real training image against the \emph{other} training images gives
0.0953 (mean nearest-neighbor cosine 0.9047), and 9.96\% of training images have at least one other training
image at cosine 0.95 or higher, so this domain is dense and FSMS cannot approach 1. Real images
that were never seen in training are the natural reference for ``not a copy'': the 1{,}512 held-out
test images score 0.1147 against the training set (3.5\% of them have a neighbor at 0.95 or
above, none at 0.97 or above).

\begin{table}[t]
\centering
\caption{Reference values of FSMS. The mean cosine is the mean over images of the cosine similarity to the
nearest training image, so FSMS is one minus it. The held-out test images are the 1{,}512 official test
images, and the last three columns give the percentage of images with a neighbor at or above the cosine
value and the highest cosine found.}
\label{tab:fsms-reference}
\small
\begin{tabular}{@{}lrrrrr@{}}
\toprule
Set & FSMS & Mean cosine & $\geq0.95$ (\%) & $\geq0.97$ (\%) & Max \\
\midrule
Training images against themselves & 0 & 1.0000 & --- & --- & --- \\
Training images against the other training images & 0.0953 & 0.9047 & 9.96 & --- & 1.000 \\
Held-out test images against the training images & 0.1147 & 0.8853 & 3.51 & 0.00 & 0.968 \\
\bottomrule

\end{tabular}
\end{table}

\paragraph{Pool audit.}
Table~\ref{tab:fsms-pool} gives FSMS for each of the 16 generation configurations. Later
checkpoints and lower CFG scales tend to lie closer to the training set. The configuration that FID-Inception ranks best (25k kimg, guidance 1.0) has the lowest FSMS of all 16 (0.1225), so selecting
by FID picks the configuration nearest the training images. Over the whole pool, 754 images
(0.27\%) have a training neighbor at cosine 0.95 or above and 7 at 0.97 or above, and the
maximum is 0.981.

\begin{table}[t]
\centering
\caption{FSMS of the synthetic pool by generation configuration (higher is safer). The bold entry
is the configuration ranked best by FID-Inception.}
\label{tab:fsms-pool}
\small
\begin{tabular}{@{}lrrrr@{}}
\toprule
Checkpoint & CFG 1.0 & CFG 1.5 & CFG 2.0 & CFG 3.0 \\
\midrule
3k & 0.2097 & 0.2065 & 0.2063 & 0.2180 \\
8k & 0.1549 & 0.1651 & 0.1766 & 0.2058 \\
15k & 0.1291 & 0.1357 & 0.1544 & 0.2070 \\
25k & \textbf{0.1225} & 0.1304 & 0.1609 & 0.2314 \\
\bottomrule

\end{tabular}
\end{table}

\paragraph{Does FSMS detect memorization?}
To test whether FSMS detects memorization, we planted copies of training images in a curated set of
8{,}012 synthetic images (FSMS 0.1318). A fixed fraction of the set, 1\%, 5\%, 10\% or 25\% (80 to 2{,}003
images), was replaced by real training images. An exact copy has a cosine similarity of 1 with its
training image, which is the highest similarity a memorized generation could have. A noisy copy, in
which Gaussian noise with a norm of 30\% of the embedding norm was added to the DINOv2 embedding,
represents a reproduction that is close to a training image but not identical. FSMS was recomputed for
each fraction and each kind of copy (Table~\ref{tab:fsms-detection}). If FSMS responds to memorization,
it should fall as the planted fraction grows, and the per-image similarity $s(x)$ of
Section~\ref{sec:methods-fsms} should be higher for the planted images than for the clean ones. FSMS falls in proportion to the planted fraction
(0.0013 for 1\%, 0.0330 for 25\%). The per-image similarity $s(x)$ also identifies the planted images:
the area under the receiver operating characteristic curve (AUROC), which is the probability that a randomly chosen planted image has a higher $s(x)$ than a
randomly chosen clean image, is 1.0 for exact copies and 0.996 for noisy copies. One percent contamination
(80 images) is a drop of about 3 to 4 seed-to-seed standard deviations of a cohort (0.0003 to 0.0004), so FSMS
resolves contamination of roughly that size. The perturbation is applied in feature space, so
this test shows the sensitivity of the statistic and not of the full image-to-feature path.

\begin{table}[t]
\centering
\caption{FSMS drop and per-image AUROC when real training images are planted into a curated set
of 8{,}012 synthetic images. Contamination is shown as a fraction of the set (number of planted
images in parentheses). The noisy copies carry feature-space Gaussian noise whose norm is 30\% of the embedding norm.}
\label{tab:fsms-detection}
\small
\begin{tabular}{@{}lrrrr@{}}
\toprule
 & \multicolumn{2}{c}{Exact copies} & \multicolumn{2}{c}{Noisy copies} \\
\cmidrule(lr){2-3}\cmidrule(lr){4-5}
Contamination & FSMS drop & AUROC & FSMS drop & AUROC \\
\midrule
1\% (80) & 0.0013 & 1.000 & 0.0009 & 0.996 \\
5\% (401) & 0.0066 & 1.000 & 0.0045 & 0.996 \\
10\% (801) & 0.0132 & 1.000 & 0.0090 & 0.996 \\
25\% (2003) & 0.0330 & 1.000 & 0.0228 & 0.996 \\
\bottomrule

\end{tabular}
\end{table}

\paragraph{Does a high similarity reflect copying or class typicality?}
DINOv2 is a general semantic embedding and not a copy detector, and the real data are two thirds
nevi, so a high cosine could mean ``a typical lesion'' and not ``close to one training image''. To
check, the DINOv2 similarity of each pool image to its nearest real image was compared with SSIM, PSNR
and LPIPS, which were computed against the same nearest real image but never use the DINOv2 embedding
(Table~\ref{tab:fsms-crossmodal}). Across all 275{,}472 pool images the DINOv2 similarity correlates with
SSIM (Pearson $r=0.42$), with PSNR ($r=0.40$) and, since LPIPS is a distance, negatively with LPIPS ($r=-0.54$).
All three correlations have $p<10^{-300}$. Across the five similarity brackets of the pool, PSNR rises and LPIPS falls at every
step, from 15.2\,dB and 0.458 in the least similar half to 22.0\,dB and 0.240 in the most similar 0.1\%.
SSIM rises up to the 99th percentile and levels off above it. High similarity therefore corresponds to
visual proximity to a specific real image.

\begin{table}[t]
\centering
\caption{Pixel-level agreement with the DINOv2 similarity. The 275{,}472 pool images are grouped by the
cosine similarity to their nearest real training image, and SSIM, PSNR and LPIPS, computed against the
same nearest real image, are averaged within each bracket. The last row gives the Pearson correlation
over all pool images (all $p<10^{-300}$). An arrow marks the direction in which the images are closer,
so LPIPS, a distance, falls as the cosine rises.}
\label{tab:fsms-crossmodal}
\small
\begin{tabular}{@{}lrrrrr@{}}
\toprule
Similarity bracket & Images & Mean cosine & SSIM $\uparrow$ & PSNR (dB) $\uparrow$ & LPIPS $\downarrow$ \\
\midrule
Least similar 50\% & 137,736 & 0.768 & 0.505 & 15.2 & 0.458 \\
50th to 90th percentile & 110,188 & 0.870 & 0.587 & 17.3 & 0.381 \\
90th to 99th percentile & 24,793 & 0.921 & 0.618 & 18.3 & 0.325 \\
99th to 99.9th percentile & 2,479 & 0.946 & 0.629 & 19.6 & 0.283 \\
Most similar 0.1\% & 276 & 0.960 & 0.628 & 22.0 & 0.240 \\
\midrule
Correlation with the cosine ($r$) & & & 0.42 & 0.40 & $-$0.54 \\
\bottomrule

\end{tabular}
\end{table}

\subsection{Generalization to ResNet-50}
\label{sec:resnet}

To test whether the findings carry over to a different architecture, the tournament, the weight
derivation, the temperature sweep and the benchmark were repeated with ResNet-50 \citep{He2016}. The
classifier settings, the pool, the metrics and the test set are the same, and the algorithm uses its own
weights derived from the ResNet-50 tournament. Two parts of the Swin-B pipeline were not repeated: the
safety sweep, so the algorithm is run at $\tau=0.97$ only (the lowest tested value with precision above
95\%), and the development of the algorithm's design, so the configuration is the one fixed in
Section~\ref{sec:algorithm}. The random control uses 30 seeds (42 to 71), which also anchors the weights.
The result is mixed. The ranking of metrics largely carries over, the per-class pattern partly does, and
the benefit of the algorithm does not. The tables behind this subsection are in
Appendix~\ref{app:resnet}.

\begin{figure}[tbp]
\centering
\includegraphics[width=0.62\linewidth]{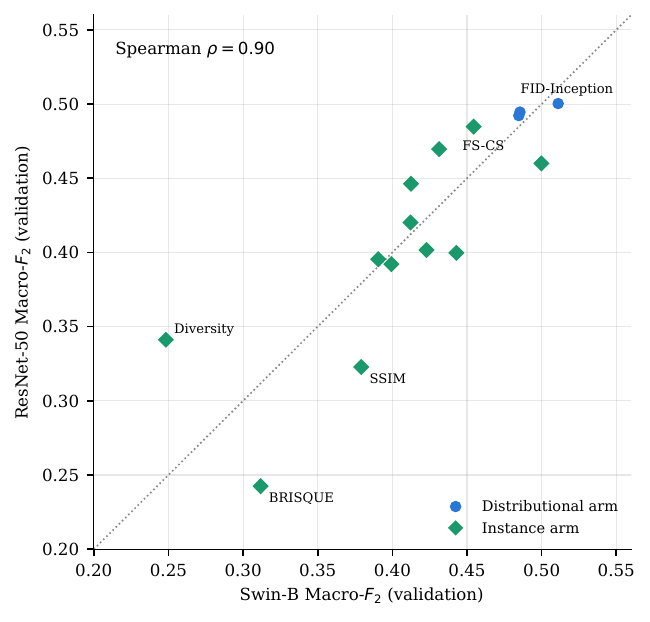}
\caption{Validation Macro-$F_2$ of each tournament arm on Swin-B and on ResNet-50. FID-DINOv2 and
KID-DINOv2 select the same configuration and are shown as one arm. The dotted line marks equal
scores.}
\label{fig:arch-agreement}
\end{figure}

\paragraph{Metric ranking.}
FID-Inception is again the strongest synthetic arm on ResNet-50 (Macro-$F_2$ 0.500, against 0.460 for
random sampling and 0.571 for real data only), and all three distinct distributional arms beat random
sampling over the 30 seeds (by $+0.041$, $+0.035$ and $+0.033$, all $p<0.001$). Among instance
metrics, Realism (0.485) and CLIP log-odds (0.470) are above random, FS-CS matches it, and the remaining
nine fall below it. Figure~\ref{fig:arch-agreement} compares the two architectures. The Macro-$F_2$ of the
15 distinct arms agrees closely (Spearman $\rho=0.90$), so a metric that serves Swin-B well on average
generally serves ResNet-50 well.

\paragraph{Per-class pattern.}
The per-class pattern agrees less well. No metric serves every class on ResNet-50 either: a different arm
leads each of the seven classes. Only AKIEC has the same leader as on Swin-B (FSIM, 0.599 against 0.596),
and the per-class rank correlation of arms between the architectures ranges from 0.37 (BKL) to 0.74
(VASC). DF is again the weakest class for FID-Inception (0.059), but the leader there is CLIP log-odds
(0.300), not FS-CS. Instance arms are single deterministic datasets, and NV is a near-tie among several arms (0.9292 for
Diversity and CLIP similarity to the artifact prompt), so part of this disagreement is noise. The
conclusion that no single metric serves every class holds on both architectures, whereas the particular
metric that serves a class does not transfer. The full per-class scores are in
Table~\ref{tab:tournament-per-class-resnet} of Appendix~\ref{app:tournament}.

\paragraph{Weights.}
The weights derived from the ResNet-50 tournament differ from the Swin-B weights. The top metric is the
same in three classes (FID-Inception for BCC, MEL and VASC) but its weight is smaller, for example 15\%
for VASC against 35\% on Swin-B. The other classes change leader. DF puts its largest weight on CLIP
log-odds (22\%) where Swin-B concentrates 33\% on SWD-DINOv2, and the distributional share ranges from
21\% to 31\% instead of 22\% to 40\%. These differences follow from the per-class leaders above, and they
show why the weights are derived per architecture. The distributional share and the
largest weight of every class are in Table~\ref{tab:resnet-weights} of Appendix~\ref{app:resnet}.

\paragraph{Temperature.}
The temperature was swept at seven values with the locked configuration. Macro-$F_2$ peaks at $T=0.05$
(0.5054) and is lower on both sides (0.4913 at $T=0.1$ and 0.4776 at $T=0.02$), and both neighboring
differences are significant ($p=0.025$ and $p=0.0004$). The maximum is therefore at the same temperature
as on Swin-B, and $T=0.05$ is used for both architectures.

\paragraph{Benchmark.}
The algorithm shows no benefit on ResNet-50. At $N=8{,}012$ and $\tau=0.97$ it scores 0.5520, against
0.5577 for the best-metric cohort and 0.5466 for random sampling, and neither difference is significant
in Welch's $t$-test ($p=0.35$ and $p=0.33$). At $N=20{,}000$ the algorithm and random sampling are also
indistinguishable ($p=0.69$). The best-metric cohort is slightly above random sampling, by 0.0112
($p=0.063$). The real-data-only run (0.5578) lies above every synthetic
cohort mean, and only 9 to 17 of the 30 runs in a synthetic cohort exceed it, so on this
architecture synthetic augmentation shows no clear gain over the real data alone.

\paragraph{What generalizes.}
Three findings carry over in part. First, the metric ranking is stable across architectures
(Spearman $\rho=0.90$), with FID-Inception strongest among synthetic arms on both. Second, no
single metric serves every class on either architecture. Third, the per-class weights are
architecture-specific, so the tournament must be repeated for each classifier. The benefit of the algorithm
does not carry over. We did not investigate why, and the per-class disagreement above, together with
the missing spread of the instance arms, makes it unclear how far the weights derived on ResNet-50 can be
trusted. We therefore claim the algorithm's benefit for Swin-B only (Section~\ref{sec:limitations}).

\section{Synthetic Dataset and Open-Source Code}
\label{sec:dataset-release}

The tournament, the algorithm and the benchmark all rest on one table: 275,472 synthetic dermoscopy
images, each with a vector of per-image scores. We release this pool and the table. Anyone can
then train classifiers on it, build a different curation method on the same scores, or repeat the
tournament with other metrics without regenerating a single image.

\subsection{Contents}
\label{sec:dataset-contents}

The release contains the 275,472 class-conditional $512 \times 512$ images described in
Section~\ref{sec:tournament-synthetic-pool}, with their composition in
Table~\ref{tab:pool-composition}: 16 generation configurations, each holding the same seven-class
mix. The grid deliberately includes under-trained and over-trained checkpoints and a wide range of
guidance, so the pool spans a broad range of image quality and of distance to the training data
(FSMS from 0.12 to 0.23 across configurations, Table~\ref{tab:fsms-pool}).

\subsection{Per-Image Metric Scores}
\label{sec:dataset-per-image-metric-scores}

Each image carries the fields in Table~\ref{tab:dataset-fields}. Twelve of the scores are the
instance metrics of the tournament (Table~\ref{tab:portfolio}). Two more, classifier confidence
and the geometric precision-recall score, were computed for the pool but are not part of the
16-metric tournament, so no downstream utility measurement accompanies them. The four
distributional metrics cannot score a single image, so they are released as one row per
generation configuration (FID with Inception and DINOv2 features, KID and SWD with DINOv2 features),
computed against the real training set. All reference-based scores use the 8{,}012 real training
images, so users who change the real reference set must recompute them. The nearest-real
similarity is the per-image term of FSMS (Section~\ref{sec:methods-fsms}), so users can filter the pool at
any similarity threshold.

\begin{table}[t]
\centering
\caption{Fields of the released per-image table (one row per image).}
\label{tab:dataset-fields}
\footnotesize
\begin{tabular}{@{}p{5.9cm}p{10.0cm}@{}}
\toprule
Field & Description \\
\midrule
\texttt{filename}, \texttt{archive} & Path of the image inside its zip archive, for example \texttt{ckpt\_25k\_cfg\_1.0/\allowbreak class\_6/\allowbreak 012418.png}, and the name of that archive (\texttt{ckpt\_25k\_cfg\_1.0.zip}). The filename is unique across the release \\
\texttt{seed} & Sampling seed of the image, which is also the number in the file name. Seeds repeat across classes and configurations \\
\texttt{source\_checkpoint\_kimg}, \texttt{cfg\_scale} & Generation configuration \\
\texttt{target\_class\_label} & Class the image was conditioned on \\
\midrule
\texttt{psnr\_score}, \texttt{ssim\_score}, \texttt{fsim\_score}, \texttt{lpips\_score} & Full-reference scores against the nearest real training image in DINOv2 space \\
\texttt{brisqe\_score} & No-reference natural-scene-statistics score \\
\texttt{fscs\_score} & DINOv2 cosine similarity to the real class centroid \\
\texttt{clip\_similarity\_cancer}, \texttt{clip\_similarity\_artifact}, \texttt{clip\_log\_odds\_score} & CLIP prompt similarities and their log ratio \\
\texttt{is\_contribution\_score} & Per-image KL term of the IS \\
\texttt{realism\_score}, \texttt{diversity\_score} & Distance to the 3rd nearest real image, and to nearby synthetic images, in DINOv2 space \\
\texttt{classifier\_confidence\_score} & Softmax probability that a ResNet-50 trained on the real training set assigns to the conditioned class (not in the tournament) \\
\texttt{geometric\_pr\_score} & Harmonic mean of the realism (inverted) and diversity scores, each clipped at its 1st and 99th percentiles and mapped to $[0,1]$ (not in the tournament) \\
\texttt{max\_sim\_to\_real} & Cosine similarity to the nearest real training image (the FSMS term) \\
\bottomrule
\end{tabular}
\end{table}

\subsection{Intended Uses}
\label{sec:dataset-intended-uses}

The release supports four uses. First, it is a pool for augmentation experiments: the cohorts of
Section~\ref{sec:benchmark} can be rebuilt from the table, and other curation methods can be compared
with them on the same pool. Second, it is a testbed for evaluation metrics, since the 16
configurations differ strongly in quality and in distance to the training set. Third, the
per-image scores allow the tournament to be repeated with additional metrics. Fourth, the
nearest-real similarity allows memorization audits.

Three limits apply. The images are synthetic and are not suitable for clinical diagnosis or
training outside research settings. The release makes no privacy guarantee: FSMS shows proximity
to training images and is not a proof of non-reproduction (Section~\ref{sec:safety-feature-space-memorization-sco}). The release
carries no skin-tone annotation, and this paper does not evaluate the fairness of models trained
on it.

\subsection{Dataset Access and Licensing}
\label{sec:dataset-access-and-licensing}

The dataset consists of the image archive (16 zip files, one per generation configuration, 97.7~GiB in total), the per-image table and the per-configuration table. The archive and the
tables are released
under the Creative Commons Attribution-NonCommercial-ShareAlike 4.0 International license
(CC BY-NC-SA 4.0). The license follows from the two sources the images depend on. The generator was
trained on ISIC-2018 training data, which is distributed under CC BY-NC, and was built with the
EDM2 code of \citet{Karras2024}, which is distributed under CC BY-NC-SA 4.0. CC BY-NC-SA 4.0 meets
the attribution, non-commercial and share-alike terms of both. Users must credit this paper and the
ISIC-2018 and HAM10000 datasets \citep{codella2019, tschandl2018}, and may not use the release for
commercial purposes.

\subsection{Open-Source Code}
\label{sec:dataset-software}

The open-source Python package \texttt{synth-curator}
(\url{https://pypi.org/project/synth-curator/} and \url{https://github.com/Amixdu/synth-curator})
applies the algorithm to a user's own synthetic pool and classifier. Given a pool, real reference images and the user's training code, it computes the
metric scores, runs the tournament, derives the weights, curates the pool, and can then train the
classifier on the curated set. The cohorts of Section~\ref{sec:benchmark} can be rebuilt with it from
the released tables.
\section{Discussion}
\label{sec:discussion}

\subsection{Findings}
\label{sec:discussion-findings}

\paragraph{Synthetic images help on Swin-B and not on ResNet-50.}
On Swin-B every synthetic cohort, including random sampling, scores above the real-data-only classifier
(0.5154), by 0.040 for random sampling and by up to 0.071 for the algorithm. This comparison is
descriptive, since the real-data-only classifier is a single run. On ResNet-50 the real-data-only
classifier (0.5578) is at or above every synthetic cohort mean. The benefit of augmentation is therefore
specific to the classifier in this study, as the benefit of curation is.

\paragraph{The value of a metric depends on the class.}
No metric led on every class in the tournament, and the arms that led across the seven
classes were FSIM, Realism, FS-CS, FID-Inception, and FID-DINOv2 together with KID-DINOv2. FID-Inception
shows the clearest case. Its classifier is the best of all arms on VASC (0.680) and shows no
measurable advantage over random sampling on DF. Class size does not explain which arm wins, since
distributional metrics lead on both the largest class (NV) and the second smallest (VASC), while an
instance metric leads on DF, which is smaller than either. The mechanism underlying this pattern
remains unclear, and
we report it as an observation. The consequence for practice is that a single global metric cannot be
best for every class, so a curation rule that uses one metric accepts a cost on some classes
without a way to see it in advance.

FID-Inception is nonetheless the strongest single arm on Macro-$F_2$ in the tournament
(0.511 against 0.475 for random sampling on Swin-B). Reports that FID can fail to track
downstream performance \citep{wu2026} therefore do not carry over to utility in this setting. FID
performed less favorably on the safety measure, because the configuration it ranks best is the one
closest to the training
images (Section~\ref{sec:safety-feature-space-memorization-sco}). Most instance metrics produced
classifiers below random sampling. FS-CS was the only one above it on Swin-B (0.500 against 0.475). Each instance arm is one
deterministic dataset with one score, so it has no seed-to-seed spread to test against. Its score is read
against the spread of random draws (standard deviation 0.021), and the gap of FS-CS is about one standard
deviation.

\paragraph{The algorithm improves on the best-metric baseline.}
On a held-out real test set, the algorithm exceeds the FID-Inception cohort by $+0.0136$ to
$+0.0164$ Macro-$F_2$ at each of four safety thresholds, and exceeds random sampling by about
$+0.03$. This is about twice the gain that best-metric sampling achieves over random sampling
($+0.0145$). The gain comes from the combination of per-class weighting and the group-scored term, and the
experiments do not isolate which component of the final configuration contributes how much. The safety
threshold adds a second control. Tightening it raises FSMS with no measurable change in Macro-$F_2$, so a
user can choose a setting at which the algorithm exceeds the best-metric baseline on utility and on safety
together, as it does at 0.90 and 0.93 here. The best-metric baseline has no corresponding setting, since the
safety of its set is fixed by the configuration it selects.

\paragraph{The result is specific to Swin-B in this study.}
ResNet-50 shows no benefit from the algorithm over either the best-metric cohort or random sampling
(Section~\ref{sec:resnet}). The ranking of metrics by
Macro-$F_2$ agrees closely across the two architectures (Spearman $\rho=0.90$), but the metric that
leads each class mostly does not, and the weights derived for ResNet-50 differ from those for Swin-B. We
did not investigate the cause. One possible reason is that synthetic images added to the real data do not
raise the score on ResNet-50 even when drawn at random, so the pool offers little for curation to improve.
We did not test this. The claim that this paper supports is therefore limited to Swin-B with
a frozen backbone and a trained linear head, and the weights were derived separately for each
classifier, so a new classifier would call for its own weights.

\paragraph{The gain over random sampling is stable when the algorithm curates more.}
At $N=20{,}000$ the algorithm still beats random sampling by $+0.0303$ Macro-$F_2$
(Section~\ref{sec:benchmark-dose-response-at-n20000}), against $+0.0281$ at $N=8{,}012$. Neither cohort changes
significantly between the two volumes, so the gap to random sampling stays the same and the extra
synthetic images add no significant gain. The procedure takes about 7\% of the pool at $N=20{,}000$ and
about 3\% at $N=8{,}012$. With two volumes, one architecture and one threshold, the data do not
describe the shape of the dose-response curve.

\paragraph{Safety is a threshold choice.}
Tightening the safety threshold raised FSMS without a measurable change in Macro-$F_2$. At $\tau=0.97$ and
$0.99$ the filter removes 7 and 0 of the 275{,}472 pool images, and the algorithm sits marginally below the
FID cohort on FSMS (by 0.0004 or less). At the tighter thresholds, 0.90 and 0.93, it is safer than the FID
cohort.

\paragraph{For FID, the best configuration is also the closest to the training images.}
The configuration that FID-Inception ranks best (25k kimg, guidance 1.0) has the lowest FSMS of the 16
generation configurations (0.1225), and the FID cohort, which draws every image from it, scores 0.1157
against 0.1639 for random sampling (Section~\ref{sec:safety-feature-space-memorization-sco}). FID is also the
strongest single metric on utility in the tournament, so on Swin-B the signal that selects the best
training data selects the configuration nearest the training images. This is consistent with training that
moves a checkpoint toward the real distribution in Inception space also moving it toward specific real
images in DINOv2 space. We did not test this.

\paragraph{Curated sets score below random sampling on safety.}
At no tested threshold does a curated cohort reach the FSMS of random sampling. For the FID cohort the
cause is visible, since it draws from the configuration nearest the training images. For the algorithm we
have not isolated the cause, although curation toward images that resemble the real data is consistent
with it. The utility gains of curation are accompanied by a lower FSMS than random sampling, and a
threshold narrows the gap without closing it.

\subsection{Inception and DINOv2 Feature Spaces}
\label{sec:discussion-feature-spaces}

Section~\ref{sec:feature-backbone} found a clear advantage of Inception-space FID on Swin-B and no
separable difference on ResNet-50.

\paragraph{The selected configuration.}
The two arms differ in the configuration they choose. FID-Inception picks 25k kimg at guidance 1.0 and
FID-DINOv2 picks 25k at 1.5. The comparison between feature spaces is therefore a comparison between
these configurations, since the feature
space acts on the result only through the configuration it selects. The direction of the guidance preference
agrees with the settings published for EDM2 on ImageNet-512, where a guidance of 1.2 to 1.4 minimizes
FID and 1.7 to 2.0 minimizes the Fr\'echet distance on DINOv2 features (FD$_\text{DINOv2}$) \citep{Karras2024}: FID on DINOv2 features prefers
stronger guidance. Guidance varies the fidelity-diversity trade-off of the pool, and one possible reason
for the better utility of the unguided configuration is that it holds more diversity. A second possible
reason is that both classifiers are frozen ImageNet-pretrained backbones, so Inception features, which
are also trained on ImageNet, may align with what the classifier sees. We tested neither.

\paragraph{Relation to earlier work.}
Stein et al.\ recommend DINOv2 features after finding that FID's Inception features track human
judgment of realism poorly \citep{stein2023}, and Woodland et al.\ find that ImageNet-based extractors rank
medical generators more consistently with expert judgment than extractors trained on medical images
\citep{woodland2024}. The criterion here differs from both, because the metrics are compared by the
utility of the classifier trained on the images they select. By that criterion FID on Inception features
did better on Swin-B, and the two cannot be separated on ResNet-50.

\paragraph{Safety.}
The configuration picked by FID-Inception has the lowest FSMS in the pool (0.1225). The configuration picked
by FID-DINOv2 has a higher FSMS (0.1304; Table~\ref{tab:fsms-pool}), so the feature space also shifts the
safety of the selected set.

\subsection{Practical Guidance}
\label{sec:discussion-practical-guidance}

The algorithm needs three things: a pool of synthetic images with per-image scores, a classifier
that can be trained many times, and a held-out real set. The procedure is
\begin{enumerate}
\item generate a pool that varies in quality and in distance to the training data, for example
across training checkpoints and CFG scales, as in Section~\ref{sec:dataset-release};
\item score the pool with a portfolio of metrics, including distributional metrics computed per
group of the pool (here, per generation configuration);
\item run the tournament on the classifier of interest and keep per-image predictions, because
per-class $F_2$ cannot be recovered from a macro-averaged table;
\item derive per-class weights (Section~\ref{sec:algorithm-per-class-utility-weights}) and
curate to a budget $N$, which is a user choice (we used the size of the real training set so that
cohorts compare at equal volume, and a surplus pool beyond that did not help);
\item choose $\tau$ from the privacy requirement, using a calibration curve built from labeled
near-duplicates if the data provide one, and confirm the result on a held-out real set. Where
privacy matters more than the last 0.01 of Macro-$F_2$, tighter thresholds are available at no
measured utility cost within the tested range.
\end{enumerate}

\subsection{Clinical Implications}
\label{sec:discussion-clinical}

Synthetic images address the scarcity of real examples for rare diagnoses described in
Section~\ref{sec:intro}, and the results show that which synthetic images are added matters for those classes. A single global quality
score such as FID did not serve every class, so a team augmenting a rare class should measure the
effect per class on held-out real data before adopting a selection rule. On Swin-B, the algorithm
raised Macro-$F_2$, which weights recall over precision and so reflects the cost of a missed lesion,
by 0.0136 to 0.0164 over the FID baseline (corrected $p<0.02$). The gain is statistically significant but modest in absolute terms, and it was
not seen on ResNet-50. The safety threshold offers a documented way to limit how closely curated
images resemble patient training images, which is relevant under regulations such as GDPR and HIPAA,
but FSMS measures proximity and is not a privacy guarantee. These results are evidence for how to
select synthetic training data in research settings. No clinician assessed the images, and the
classifiers are not validated for clinical use.

\subsection{Reusable Artifacts}
\label{sec:discussion-reusable-artifacts}

Three artifacts are useful without adopting the algorithm. The tournament table
(Figures~\ref{fig:tournament-ranking} and~\ref{fig:tournament-heatmap}) gives per-class utility
for 16 metrics on a fixed pool, which lets a practitioner pick a metric for a class of interest.
The released pool and per-image scores (Section~\ref{sec:dataset-release}) allow other curation methods,
other metrics and other classifiers to be tested against the same data. The released package \texttt{synth-curator} (Section~\ref{sec:dataset-software})
takes any pool, any metric portfolio and any classifier as input, so the tournament can be repeated
on other datasets. It ships weights precomputed for ISIC-2018 on Swin-B and ResNet-50 so that the
tournament can be skipped.

\section{Limitations}
\label{sec:limitations}

\paragraph{One dataset, one generator, one training regime.}
All results use ISIC-2018 and a single generator family (EDM2). The held-out test set comes from the
same challenge as the training data and is not an external dataset, so we do not know how the
results transfer to other sites, devices or generators. Every classifier is a frozen
ImageNet-pretrained backbone with a trained linear head. The frozen backbone is a deliberate choice
(Section~\ref{sec:tournament-tournament-protocol}) that isolates the effect of the selected images. The cost is external validity:
deployed classifiers are usually fine-tuned, which can use synthetic images differently, so the
tournament weights and the benchmark gains may differ.

\paragraph{Evidence on one architecture.}
The benefit of the algorithm is shown for Swin-B. ResNet-50
shows none (Section~\ref{sec:resnet}), and we did not investigate why. Metric rankings agree across
the two architectures, but the per-class leaders do not, so the per-class weights are specific to the
classifier. With two architectures we cannot say whether the Swin-B result or the ResNet-50 result is more
typical of other classifiers.

\paragraph{Tournament noise and scope.}
The tournament is scored on a validation split that also selects the early-stopping epoch, so
absolute scores are optimistic, and the validation split has only 23 DF and 28 VASC images. The
twelve instance arms are single deterministic datasets with one score each and no seed-to-seed spread, so their per-class values are indicative and a gap of 0.01
to 0.02 to random is inside the spread of random draws. The distributional arms use 30 seeds but
select only three distinct generation configurations. All four beat random sampling, which is also
what concentrating on any single configuration might produce. A matched control, a randomly
chosen single configuration, was never run, so the tournament cannot fully separate a metric
choosing a good configuration from one configuration beating the mixture.

\paragraph{Sensitivity of the weights.}
The weights are derived from these noisy measurements through a hard threshold at the random
baseline. A metric near that baseline can be included or excluded by baseline noise alone. The worst-metric anchor is also portfolio-dependent, so weights are not comparable across
portfolios, and reproducing them requires the same 15-metric portfolio. We did not
test a margin criterion or replicated instance arms.

\paragraph{Exploratory development.}
The temperature of 0.05 was the best of three values without a significant
difference to its neighbors. The mechanism by which the group-scored term helps VASC was not
demonstrated, and VASC remains below the FID baseline on the test set. No control with equal or
mismatched weights was run, so the contribution of the per-class weights cannot be separated from that
of the other components.
\paragraph{Benchmark scope.}
The test set has 1{,}512 images, and AKIEC, DF and VASC have only 35 to 44 images each, so
per-class scores are noisy and the per-class tests are uncorrected. Classifier initialization is
fixed, so cohort spread reflects the selected data and not training variance. The fixed seed also
does not make runs bit-reproducible, since PyTorch's full deterministic mode was not enabled and
some GPU matrix operations can differ between executions. In the tournament, two arms with
identical curated sets produced identical predictions in all 60 matched pairs, so this variation
was not visible there. The real-data-only baseline is a single run and cannot be tested. The
algorithm was evaluated at two budgets, $N=8{,}012$ and $N=20{,}000$, both at or above the real
training set size, and the $N=20{,}000$ cohort and the ResNet-50 runs exist at one safety threshold
($\tau=0.97$). Behavior at much smaller or much larger budgets is untested. We did not measure the
compute needed to generate a surplus pool and curate it, against generating exactly $N$ images.

\paragraph{What FSMS does and does not show.}
FSMS is a mean over a set, whereas privacy risk concentrates in the closest images, and we report
counts of pool images above fixed similarity values. Appendix~\ref{app:benchmark-tables} reports the
maximum, the 99.9th percentile and the share of images above fixed similarity values for each cohort.
Similarity is measured in one embedding space (DINOv2). The calibration of $\tau$ uses
re-photographs of the same lesion as positives. These differ from diffusion reproductions, so the
calibration bounds false removals and says nothing about missed reproductions. At $\tau=0.97$ the
filter identifies about 10\% of same-lesion pairs and removes 7 of 275{,}472 pool images. All
curated cohorts score below random sampling on FSMS at every tested threshold. FSMS shows proximity
to training images and is not a privacy guarantee.

\paragraph{Fairness and clinical use.}
We do not evaluate whether curation or augmentation changes performance across skin tones. The
synthetic pool carries no skin-tone labels, and no clinician assessed the images. The pool and the
trained models are research artifacts and are not validated for clinical use.

\section{Conclusion}
\label{sec:conclusion}

We measured what 16 evaluation metrics are worth for curating synthetic dermoscopy images, class
by class, by training classifiers on the images each metric selects. No single metric served every
class. FID-Inception, the usual choice, produced the best classifier on one class and no measurable
advantage over random sampling on another. We turned these per-class measurements into a safety-aware
evaluation algorithm with per-class weights and a safety threshold calibrated on labeled
near-duplicates of real images.

On a held-out real test set, the algorithm improves Macro-$F_2$ over the FID-Inception baseline for
Swin-B by $+0.0136$ to $+0.0164$ across four safety thresholds, significant after correction at each, and over random sampling by $+0.0281$ to $+0.0309$. Tightening the safety threshold raised FSMS without a
measurable change in Macro-$F_2$, but every curated cohort remains less safe than random sampling.

On the primary classifier, Swin-B, our proposed algorithm improved upon FID-based selection by a modest
but statistically significant margin, offering a practical, safety-aware procedure for choosing synthetic
training images. We release the 275{,}472-image pool with its per-image scores and the open-source package
\texttt{synth-curator}, enabling others to test alternative curation methods and metrics on the same data
and to apply this algorithm to other pools and classifiers.

\section*{Data and Code Availability}

The 275{,}472 synthetic images, the per-image metric scores and the per-configuration table are
available on Zenodo at \url{https://doi.org/10.5281/zenodo.23851497} under CC BY-NC-SA 4.0. 
The open-source Python package \texttt{synth-curator} (\url{https://github.com/Amixdu/synth-curator})
enables researchers to apply this curation algorithm to their own synthetic datasets and custom
classifier architectures. The raw result tables behind every figure and table, the
scripts that produce the reported numbers and the code of the pipeline stages used in this study are
available at \url{https://github.com/Amixdu/synth-curator-paper}.

\section*{Acknowledgements}

This work was supported by computational resources and services provided by the Bunya
supercomputer of the University of Queensland Research Computing Centre \citep{rcc2024bunya} and by
the Crux GPU server of Southern Cross University. The authors thank the administrators of both
systems for their technical support.

\section*{Use of AI Tools}

During the preparation of this work, the authors used Artificial Intelligence tools (Claude by
Anthropic and Gemini by Google) to improve the language, readability, debugging and formatting of the
manuscript. Additionally, these tools were used to assist in formatting, refactoring, debugging, and 
documenting the associated open-source code repositories. After using these tools, the authors 
thoroughly reviewed and edited the content, and take
full responsibility for the final publication.

\clearpage
\bibliographystyle{plainnat}
\bibliography{references}

\clearpage
\section*{Appendix}
\appendix
\renewcommand{\thetable}{A\arabic{table}}
\renewcommand{\thefigure}{A\arabic{figure}}
\renewcommand{\theHtable}{A\arabic{table}}
\renewcommand{\theHfigure}{A\arabic{figure}}
\setcounter{table}{0}
\setcounter{figure}{0}
\section{Full Cohort Results}
\label{app:cohorts}

Tables~\ref{tab:cohort-metrics-swin} and~\ref{tab:cohort-metrics-resnet} give the test-set metrics of
every cohort in Section~\ref{sec:benchmark}: Macro-$F_2$, Macro-$F_1$, macro-averaged ROC-AUC, macro
recall, macro precision and overall accuracy. Entries are mean $\pm$ standard deviation over 30
curation seeds, and the real-only row is a single run.

\begin{table}[t]
\centering
\caption{Swin-B test-set metrics for every cohort. $\tau$ is the safety threshold.}
\label{tab:cohort-metrics-swin}
\footnotesize
\setlength{\tabcolsep}{3pt}
\begin{tabular}{@{}lrllllll@{}}
\toprule
Cohort & $N$ & Macro-$F_2$ & Macro-$F_1$ & ROC-AUC & Recall & Precision & Accuracy \\
\midrule
Real data only & --- & 0.515 & 0.562 & 0.937 & 0.493 & 0.716 & 0.751 \\
\midrule
Random & $8{,}012$ & 0.556 $\pm$ 0.022 & 0.592 $\pm$ 0.020 & 0.935 $\pm$ 0.002 & 0.537 $\pm$ 0.023 & 0.697 $\pm$ 0.027 & 0.761 $\pm$ 0.007 \\
Best-metric (FID-Inception) & $8{,}012$ & 0.570 $\pm$ 0.020 & 0.604 $\pm$ 0.017 & 0.938 $\pm$ 0.002 & 0.552 $\pm$ 0.022 & 0.698 $\pm$ 0.020 & 0.767 $\pm$ 0.005 \\
\midrule
Algorithm, $\tau=0.90$ & $8{,}012$ & 0.586 $\pm$ 0.016 & 0.619 $\pm$ 0.014 & 0.939 $\pm$ 0.001 & 0.568 $\pm$ 0.017 & 0.708 $\pm$ 0.023 & 0.773 $\pm$ 0.005 \\
Algorithm, $\tau=0.93$ & $8{,}012$ & 0.587 $\pm$ 0.018 & 0.620 $\pm$ 0.018 & 0.939 $\pm$ 0.002 & 0.569 $\pm$ 0.019 & 0.709 $\pm$ 0.024 & 0.773 $\pm$ 0.008 \\
Algorithm, $\tau=0.97$ & $8{,}012$ & 0.584 $\pm$ 0.014 & 0.616 $\pm$ 0.012 & 0.939 $\pm$ 0.002 & 0.567 $\pm$ 0.015 & 0.701 $\pm$ 0.020 & 0.771 $\pm$ 0.004 \\
Algorithm, $\tau=0.99$ & $8{,}012$ & 0.584 $\pm$ 0.013 & 0.617 $\pm$ 0.013 & 0.939 $\pm$ 0.001 & 0.567 $\pm$ 0.013 & 0.701 $\pm$ 0.019 & 0.771 $\pm$ 0.005 \\
\midrule
Random & $20{,}000$ & 0.555 $\pm$ 0.022 & 0.591 $\pm$ 0.018 & 0.932 $\pm$ 0.002 & 0.538 $\pm$ 0.024 & 0.703 $\pm$ 0.020 & 0.758 $\pm$ 0.008 \\
Algorithm, $\tau=0.97$ & $20{,}000$ & 0.586 $\pm$ 0.016 & 0.618 $\pm$ 0.013 & 0.937 $\pm$ 0.002 & 0.569 $\pm$ 0.019 & 0.712 $\pm$ 0.021 & 0.771 $\pm$ 0.004 \\
\bottomrule

\end{tabular}
\end{table}

\begin{table}[t]
\centering
\caption{ResNet-50 test-set metrics for every cohort.}
\label{tab:cohort-metrics-resnet}
\footnotesize
\setlength{\tabcolsep}{3pt}
\begin{tabular}{@{}lrllllll@{}}
\toprule
Cohort & $N$ & Macro-$F_2$ & Macro-$F_1$ & ROC-AUC & Recall & Precision & Accuracy \\
\midrule
Real data only & --- & 0.558 & 0.583 & 0.931 & 0.550 & 0.714 & 0.763 \\
\midrule
Random & $8{,}012$ & 0.547 $\pm$ 0.020 & 0.578 $\pm$ 0.019 & 0.933 $\pm$ 0.003 & 0.533 $\pm$ 0.021 & 0.693 $\pm$ 0.027 & 0.764 $\pm$ 0.004 \\
Best-metric (FID-Inception) & $8{,}012$ & 0.558 $\pm$ 0.025 & 0.587 $\pm$ 0.023 & 0.936 $\pm$ 0.003 & 0.544 $\pm$ 0.025 & 0.687 $\pm$ 0.023 & 0.767 $\pm$ 0.006 \\
\midrule
Algorithm, $\tau=0.97$ & $8{,}012$ & 0.552 $\pm$ 0.022 & 0.583 $\pm$ 0.021 & 0.935 $\pm$ 0.002 & 0.538 $\pm$ 0.022 & 0.696 $\pm$ 0.021 & 0.766 $\pm$ 0.007 \\
\midrule
Random & $20{,}000$ & 0.556 $\pm$ 0.024 & 0.578 $\pm$ 0.022 & 0.930 $\pm$ 0.002 & 0.548 $\pm$ 0.026 & 0.665 $\pm$ 0.028 & 0.761 $\pm$ 0.006 \\
Algorithm, $\tau=0.97$ & $20{,}000$ & 0.554 $\pm$ 0.021 & 0.582 $\pm$ 0.018 & 0.932 $\pm$ 0.003 & 0.541 $\pm$ 0.023 & 0.682 $\pm$ 0.028 & 0.761 $\pm$ 0.007 \\
\bottomrule

\end{tabular}
\end{table}

\section{Tournament Results in Full}
\label{app:tournament}

Tables~\ref{tab:tournament-overall} and~\ref{tab:tournament-overall-resnet} give the Macro-$F_2$ of every
arm for Swin-B and ResNet-50, behind Figure~\ref{fig:tournament-ranking} and
Section~\ref{sec:resnet}. Tables~\ref{tab:tournament-per-class} and~\ref{tab:tournament-per-class-resnet}
give the per-class $F_2$ of the two architectures, behind Figure~\ref{fig:tournament-heatmap}.

\begin{table}[t]
\centering
\caption{Swin-B Macro-$F_2$ on the validation split for synthetic-only training. Distributional and
random-control entries are mean $\pm$ standard deviation over seeds, instance entries are single
deterministic datasets. $\Delta$ is the difference from the mean of the random control (60 runs); comparisons
in the text use Welch's $t$-test against all 60 random runs. FID-DINOv2 and KID-DINOv2 select the same configuration and
have identical results.}
\label{tab:tournament-overall}
\small
\begin{tabular}{@{}llrrr@{}}
\toprule
Metric & Family & Runs & Macro-$F_2$ & $\Delta$ \\
\midrule
FID-Inception & Distributional & 30 & 0.511 $\pm$ 0.015 & +0.036 \\
SWD-DINOv2 & Distributional & 30 & 0.485 $\pm$ 0.017 & +0.011 \\
FID-DINOv2 & Distributional & 30 & 0.485 $\pm$ 0.018 & +0.010 \\
KID-DINOv2 & Distributional & 30 & 0.485 $\pm$ 0.018 & +0.010 \\
\midrule
FS-CS & Instance & 1 & 0.500 & +0.025 \\
Realism & Instance & 1 & 0.454 & $-$0.020 \\
CLIP sim. (artifact) & Instance & 1 & 0.443 & $-$0.032 \\
CLIP log-odds & Instance & 1 & 0.431 & $-$0.043 \\
LPIPS & Instance & 1 & 0.423 & $-$0.052 \\
FSIM & Instance & 1 & 0.412 & $-$0.062 \\
PSNR & Instance & 1 & 0.412 & $-$0.063 \\
CLIP sim. (cancer) & Instance & 1 & 0.399 & $-$0.075 \\
IS contribution & Instance & 1 & 0.391 & $-$0.084 \\
SSIM & Instance & 1 & 0.379 & $-$0.096 \\
BRISQUE & Instance & 1 & 0.312 & $-$0.163 \\
Diversity & Instance & 1 & 0.248 & $-$0.227 \\
\midrule
Random control & Baseline & 60 & 0.475 $\pm$ 0.021 & --- \\
Real data only & Reference & 1 & 0.542 & +0.067 \\
\bottomrule

\end{tabular}
\end{table}

\begin{table}[t]
\centering
\caption{ResNet-50 Macro-$F_2$ on the validation split for synthetic-only training, in the same
format as Table~\ref{tab:tournament-overall}. The random control has 30 runs.}
\label{tab:tournament-overall-resnet}
\small
\begin{tabular}{@{}llrrr@{}}
\toprule
Metric & Family & Runs & Macro-$F_2$ & $\Delta$ \\
\midrule
FID-Inception & Distributional & 30 & 0.500 $\pm$ 0.018 & +0.041 \\
SWD-DINOv2 & Distributional & 30 & 0.495 $\pm$ 0.026 & +0.035 \\
FID-DINOv2 & Distributional & 30 & 0.492 $\pm$ 0.027 & +0.033 \\
KID-DINOv2 & Distributional & 30 & 0.492 $\pm$ 0.027 & +0.033 \\
\midrule
Realism & Instance & 1 & 0.485 & +0.025 \\
CLIP log-odds & Instance & 1 & 0.470 & +0.010 \\
FS-CS & Instance & 1 & 0.460 & +0.000 \\
FSIM & Instance & 1 & 0.446 & $-$0.013 \\
PSNR & Instance & 1 & 0.420 & $-$0.039 \\
LPIPS & Instance & 1 & 0.402 & $-$0.058 \\
CLIP sim. (artifact) & Instance & 1 & 0.400 & $-$0.060 \\
IS contribution & Instance & 1 & 0.395 & $-$0.064 \\
CLIP sim. (cancer) & Instance & 1 & 0.392 & $-$0.067 \\
Diversity & Instance & 1 & 0.341 & $-$0.118 \\
SSIM & Instance & 1 & 0.323 & $-$0.137 \\
BRISQUE & Instance & 1 & 0.242 & $-$0.217 \\
\midrule
Random control & Baseline & 30 & 0.460 $\pm$ 0.032 & --- \\
Real data only & Reference & 1 & 0.571 & +0.112 \\
\bottomrule

\end{tabular}
\end{table}

\begin{table}[t]
\centering
\caption{Per-class $F_2$ on the validation split for Swin-B. Bold marks the best synthetic arm
in each column. The last column is Macro-$F_2$. FID-DINOv2 and KID-DINOv2 tie in the NV
column and are identical throughout.}
\label{tab:tournament-per-class}
\small
\setlength{\tabcolsep}{4pt}
\begin{tabular}{@{}lrrrrrrrr@{}}
\toprule
 & AKIEC & BCC & BKL & DF & MEL & NV & VASC & Macro \\
\midrule
Validation images & 65 & 103 & 220 & 23 & 223 & 1,341 & 28 & 2,003 \\
\midrule
FID-Inception & 0.375 & 0.583 & 0.502 & 0.137 & 0.373 & 0.927 & \textbf{0.680} & \textbf{0.511} \\
SWD-DINOv2 & 0.329 & 0.572 & 0.510 & 0.168 & 0.348 & 0.935 & 0.537 & 0.485 \\
FID-DINOv2 & 0.367 & 0.547 & 0.494 & 0.143 & 0.338 & \textbf{0.936} & 0.568 & 0.485 \\
KID-DINOv2 & 0.367 & 0.547 & 0.494 & 0.143 & 0.338 & \textbf{0.936} & 0.568 & 0.485 \\
\midrule
FS-CS & 0.515 & 0.569 & 0.494 & \textbf{0.225} & 0.395 & 0.860 & 0.441 & 0.500 \\
Realism & 0.400 & \textbf{0.597} & \textbf{0.561} & 0.053 & \textbf{0.445} & 0.807 & 0.317 & 0.454 \\
CLIP sim. (artifact) & 0.419 & 0.474 & 0.479 & 0.054 & 0.360 & 0.920 & 0.394 & 0.443 \\
CLIP log-odds & 0.332 & 0.441 & 0.421 & 0.104 & 0.287 & 0.919 & 0.515 & 0.431 \\
LPIPS & 0.537 & 0.425 & 0.451 & 0.103 & 0.268 & 0.924 & 0.252 & 0.423 \\
FSIM & \textbf{0.596} & 0.395 & 0.346 & 0.054 & 0.326 & 0.918 & 0.252 & 0.412 \\
PSNR & 0.130 & 0.539 & 0.441 & 0.105 & 0.386 & 0.924 & 0.360 & 0.412 \\
CLIP sim. (cancer) & 0.418 & 0.546 & 0.401 & 0.000 & 0.181 & 0.927 & 0.323 & 0.399 \\
IS contribution & 0.319 & 0.244 & 0.499 & 0.052 & 0.290 & 0.929 & 0.400 & 0.391 \\
SSIM & 0.141 & 0.357 & 0.498 & 0.106 & 0.407 & 0.851 & 0.294 & 0.379 \\
BRISQUE & 0.074 & 0.116 & 0.448 & 0.054 & 0.346 & 0.932 & 0.212 & 0.312 \\
Diversity & 0.094 & 0.173 & 0.222 & 0.000 & 0.148 & 0.929 & 0.171 & 0.248 \\
\midrule
Random control & 0.343 & 0.528 & 0.492 & 0.156 & 0.326 & 0.930 & 0.548 & 0.475 \\
Real data only & 0.459 & 0.631 & 0.564 & 0.288 & 0.436 & 0.937 & 0.480 & 0.542 \\
\bottomrule

\end{tabular}
\end{table}

Table~\ref{tab:tournament-per-class-resnet} gives the per-class $F_2$ for ResNet-50, the
counterpart of Table~\ref{tab:tournament-per-class}.

\begin{table}[t]
\centering
\caption{Per-class $F_2$ on the validation split for ResNet-50. Bold marks the best synthetic arm
in each column. The last column is Macro-$F_2$.}
\label{tab:tournament-per-class-resnet}
\small
\setlength{\tabcolsep}{4pt}
\begin{tabular}{@{}lrrrrrrrr@{}}
\toprule
 & AKIEC & BCC & BKL & DF & MEL & NV & VASC & Macro \\
\midrule
Validation images & 65 & 103 & 220 & 23 & 223 & 1,341 & 28 & 2,003 \\
\midrule
FID-Inception & 0.513 & \textbf{0.573} & 0.446 & 0.059 & 0.369 & 0.922 & 0.620 & \textbf{0.500} \\
SWD-DINOv2 & 0.502 & 0.554 & 0.497 & 0.111 & 0.346 & 0.922 & 0.531 & 0.495 \\
FID-DINOv2 & 0.525 & 0.543 & 0.520 & 0.123 & 0.301 & 0.925 & 0.509 & 0.492 \\
KID-DINOv2 & 0.525 & 0.543 & 0.520 & 0.123 & 0.301 & 0.925 & 0.509 & 0.492 \\
\midrule
Realism & 0.574 & 0.455 & 0.482 & 0.051 & 0.351 & 0.856 & \textbf{0.625} & 0.485 \\
CLIP log-odds & 0.531 & 0.256 & 0.460 & \textbf{0.300} & 0.205 & 0.915 & 0.621 & 0.470 \\
FS-CS & 0.422 & 0.477 & 0.447 & 0.175 & 0.255 & 0.896 & 0.547 & 0.460 \\
FSIM & \textbf{0.599} & 0.469 & 0.492 & 0.152 & 0.216 & 0.905 & 0.292 & 0.446 \\
PSNR & 0.094 & 0.488 & 0.494 & 0.103 & \textbf{0.397} & 0.897 & 0.469 & 0.420 \\
LPIPS & 0.493 & 0.380 & 0.358 & 0.105 & 0.353 & 0.909 & 0.214 & 0.402 \\
CLIP sim. (artifact) & 0.548 & 0.465 & 0.271 & 0.054 & 0.205 & 0.929 & 0.325 & 0.400 \\
IS contribution & 0.472 & 0.182 & \textbf{0.533} & 0.105 & 0.181 & 0.929 & 0.366 & 0.395 \\
CLIP sim. (cancer) & 0.553 & 0.300 & 0.371 & 0.000 & 0.228 & 0.927 & 0.366 & 0.392 \\
Diversity & 0.165 & 0.389 & 0.325 & 0.053 & 0.241 & \textbf{0.929} & 0.285 & 0.341 \\
SSIM & 0.142 & 0.225 & 0.438 & 0.000 & 0.347 & 0.898 & 0.210 & 0.323 \\
BRISQUE & 0.144 & 0.092 & 0.281 & 0.000 & 0.255 & 0.925 & 0.000 & 0.242 \\
\midrule
Random control & 0.525 & 0.471 & 0.469 & 0.085 & 0.303 & 0.926 & 0.439 & 0.460 \\
Real data only & 0.592 & 0.722 & 0.623 & 0.106 & 0.393 & 0.919 & 0.644 & 0.571 \\
\bottomrule

\end{tabular}
\end{table}

Table~\ref{tab:distributional-pool} gives the scores of the four distributional metrics for each of the 16
generation configurations. Each distributional arm of the tournament curates from the configuration that
has the best score for its metric.

\begin{table}[t]
\centering
\caption{Distributional metric scores of the 16 generation configurations against the real training set
(17,217 synthetic images each; computation in Appendix~\ref{app:metrics}). Lower is better for all four, and bold marks the best configuration
for each metric, which is the one the corresponding tournament arm curates from.}
\label{tab:distributional-pool}
\small
\begin{tabular}{@{}llrrrr@{}}
\toprule
Checkpoint & CFG & FID-Inception & FID-DINOv2 & KID-DINOv2 & SWD-DINOv2 \\
\midrule
3k & 1.0 & 120.1 & 1325.7 & 23.19 & 0.926 \\
 & 1.5 & 114.0 & 1217.4 & 20.75 & 0.822 \\
 & 2.0 & 111.2 & 1138.2 & 18.41 & 0.801 \\
 & 3.0 & 116.5 & 1064.3 & 16.07 & 0.773 \\
\midrule
8k & 1.0 & 65.1 & 652.0 & 11.20 & 0.562 \\
 & 1.5 & 57.8 & 569.3 & 9.25 & 0.532 \\
 & 2.0 & 58.7 & 533.9 & 8.13 & 0.479 \\
 & 3.0 & 75.1 & 571.5 & 8.44 & 0.516 \\
\midrule
15k & 1.0 & 36.4 & 322.3 & 5.30 & 0.373 \\
 & 1.5 & 33.4 & 258.1 & 3.93 & \textbf{0.316} \\
 & 2.0 & 47.2 & 306.0 & 4.64 & 0.354 \\
 & 3.0 & 85.7 & 522.9 & 8.02 & 0.479 \\
\midrule
25k & 1.0 & \textbf{29.7} & 250.3 & 3.97 & 0.333 \\
 & 1.5 & 35.1 & \textbf{227.3} & \textbf{3.59} & 0.319 \\
 & 2.0 & 57.6 & 332.5 & 5.36 & 0.351 \\
 & 3.0 & 99.5 & 653.0 & 10.54 & 0.594 \\
\bottomrule

\end{tabular}
\end{table}

\section{Threshold Calibration Curve}
\label{app:calibration}

Table~\ref{tab:calibration-bands} gives the calibration data behind Figure~\ref{fig:calibration}. The
pairs are the 117{,}751 pairs of real training images with cosine similarity at or above 0.90, of which
776 share a lesion identifier. Cumulative columns refer to pairs at or above the lower edge of the
band.

\begin{table}[t]
\centering
\caption{Same-lesion precision and recall by cosine band for real training pairs. ``In band'' columns
use pairs inside the band, and ``cumulative'' columns use all pairs at or above its lower edge.}
\label{tab:calibration-bands}
\footnotesize
\setlength{\tabcolsep}{4pt}
\begin{tabular}{@{}lrrrrrr@{}}
\toprule
Cosine band & Pairs & Same lesion & Precision (in band) & Pairs (cumulative) & Precision (cumulative) & Recall (cumulative) \\
\midrule
0.900--0.920 & 91,548 & 237 & 0.3\% & 117,751 & 0.7\% & 100.0\% \\
0.920--0.930 & 16,709 & 112 & 0.7\% & 26,203 & 2.1\% & 69.5\% \\
0.930--0.940 & 6,835 & 105 & 1.5\% & 9,494 & 4.5\% & 55.0\% \\
0.940--0.950 & 2,020 & 93 & 4.6\% & 2,659 & 12.1\% & 41.5\% \\
0.950--0.965 & 524 & 121 & 23.1\% & 639 & 35.8\% & 29.5\% \\
0.965--0.970 & 35 & 31 & 88.6\% & 115 & 93.9\% & 13.9\% \\
0.970--0.980 & 45 & 43 & 95.6\% & 80 & 96.2\% & 9.9\% \\
0.980--0.990 & 29 & 28 & 96.6\% & 35 & 97.1\% & 4.4\% \\
$\ge$0.990 & 6 & 6 & 100.0\% & 6 & 100.0\% & 0.8\% \\
\bottomrule

\end{tabular}
\end{table}

\section{Locked Metric Weights}
\label{app:weights}

Table~\ref{tab:weights-full} lists the weight (in percent) of each of the 15 metrics for every
class on Swin-B, from Equations~\ref{eq:weight-a} and~\ref{eq:weight-b}.

\begin{table}[t]
\centering
\caption{Locked per-class metric weights for Swin-B, in percent. A value of 0 means the metric's
weight is zero or below 0.05\%.}
\label{tab:weights-full}
\small
\setlength{\tabcolsep}{4pt}
\begin{tabular}{@{}lrrrrrrr@{}}
\toprule
Metric & AKIEC & BCC & BKL & DF & MEL & NV & VASC \\
\midrule
FID-Inc. & 12.4 & 13.1 & 8.0 & 0 & 13.4 & 0 & 35.0 \\
KID-DINO & 9.2 & 4.5 & 1.4 & 0 & 3.3 & 12.7 & 5.4 \\
SWD-DINO & 0 & 10.4 & 15.0 & 33.0 & 6.3 & 11.8 & 0 \\
\midrule
CLIP log-odds & 6.6 & 6.7 & 5.8 & 7.7 & 5.2 & 7.3 & 10.9 \\
FS-CS & 11.2 & 9.4 & 7.9 & 16.6 & 9.2 & 3.4 & 8.6 \\
CLIP cancer & 8.7 & 8.9 & 5.2 & 0 & 1.2 & 7.8 & 4.8 \\
CLIP artifact & 8.8 & 7.4 & 7.5 & 4.0 & 7.9 & 7.4 & 7.1 \\
LPIPS & 11.8 & 6.4 & 6.7 & 7.6 & 4.5 & 7.6 & 2.6 \\
FSIM & 13.3 & 5.8 & 3.6 & 4.0 & 6.7 & 7.3 & 2.6 \\
BRISQUE & 0 & 0 & 6.6 & 4.0 & 7.4 & 8.2 & 1.3 \\
PSNR & 1.4 & 8.7 & 6.4 & 7.7 & 8.9 & 7.6 & 6.0 \\
SSIM & 1.7 & 5.0 & 8.0 & 7.8 & 9.7 & 2.9 & 3.9 \\
Diversity & 0.5 & 1.2 & 0 & 0 & 0 & 8.0 & 0 \\
Realism & 8.3 & 9.9 & 9.9 & 3.9 & 11.1 & 0 & 4.6 \\
IS contr. & 6.2 & 2.7 & 8.1 & 3.8 & 5.3 & 8.0 & 7.3 \\
\midrule
Total & 100 & 100 & 100 & 100 & 100 & 100 & 100 \\
\bottomrule

\end{tabular}
\end{table}

\section{Benchmark Comparisons in Full}
\label{app:benchmark-tables}

Table~\ref{tab:benchmark-stats} gives the exact values behind Figure~\ref{fig:forest}, including the
Mann-Whitney tests. The per-class results follow. The ResNet-50 counterparts are in
Appendix~\ref{app:resnet}.

\begin{table}[t]
\centering
\caption{Swin-B test-set comparisons of Macro-$F_2$ (difference, 95\% confidence interval from
Welch's $t$-test, Cohen's $d$, Welch $p$, Bonferroni-adjusted $p$ over the four primary tests,
and Mann-Whitney $p$). Rows below the rule are secondary and uncorrected.}
\label{tab:benchmark-stats}
\footnotesize
\setlength{\tabcolsep}{3pt}
\begin{tabular}{@{}lrrrrrr@{}}
\toprule
Comparison & $\Delta$ & 95\% CI & $d$ & Welch $p$ & Adj.\ $p$ & MWU $p$ \\
\midrule
Algorithm ($\tau=0.90$) vs.\ Best-metric & +0.0156 & [+0.0062, +0.0250] & 0.86 & 0.0016 & 0.0062 & 0.0020 \\
Algorithm ($\tau=0.93$) vs.\ Best-metric & +0.0164 & [+0.0065, +0.0264] & 0.85 & 0.0016 & 0.0066 & 0.0023 \\
Algorithm ($\tau=0.97$) vs.\ Best-metric & +0.0136 & [+0.0046, +0.0226] & 0.78 & 0.0039 & 0.0155 & 0.0044 \\
Algorithm ($\tau=0.99$) vs.\ Best-metric & +0.0140 & [+0.0052, +0.0228] & 0.83 & 0.0024 & 0.0095 & 0.0011 \\
\midrule
Algorithm ($\tau=0.90$) vs.\ Random, $N=8{,}012$ & +0.0301 & [+0.0202, +0.0401] & 1.57 & $1.5\times10^{-7}$ & --- & $2.3\times10^{-6}$ \\
Algorithm ($\tau=0.93$) vs.\ Random, $N=8{,}012$ & +0.0310 & [+0.0205, +0.0415] & 1.52 & $2.2\times10^{-7}$ & --- & $2.5\times10^{-6}$ \\
Algorithm ($\tau=0.97$) vs.\ Random, $N=8{,}012$ & +0.0281 & [+0.0185, +0.0378] & 1.52 & $3.7\times10^{-7}$ & --- & $3.8\times10^{-6}$ \\
Algorithm ($\tau=0.99$) vs.\ Random, $N=8{,}012$ & +0.0286 & [+0.0191, +0.0380] & 1.58 & $2.0\times10^{-7}$ & --- & $3.1\times10^{-6}$ \\
Best-metric vs.\ Random, $N=8{,}012$ & +0.0145 & [+0.0036, +0.0255] & 0.68 & 0.0103 & --- & 0.0067 \\
Algorithm ($\tau=0.97$) vs.\ Random, $N=20{,}000$ & +0.0303 & [+0.0201, +0.0404] & 1.55 & $1.8\times10^{-7}$ & --- & $1.9\times10^{-6}$ \\
Algorithm ($\tau=0.97$), $N=20{,}000$ vs.\ $N=8{,}012$ & +0.0017 & [$-$0.0061, +0.0096] & 0.11 & 0.6584 & --- & 0.8187 \\
Random $N=20{,}000$ vs.\ $N=8{,}012$ & $-$0.0004 & [$-$0.0119, +0.0111] & $-$0.02 & 0.9440 & --- & 0.9823 \\
\bottomrule

\end{tabular}
\end{table}

Table~\ref{tab:benchmark-class-diff} gives the per-class differences from FID-Inception and from random
sampling at each threshold, and Table~\ref{tab:benchmark-per-class} lists the per-class scores. Against
FID-Inception the pattern is the same at all four thresholds. The algorithm is higher on AKIEC, BCC
and DF, lower on VASC and MEL, with no detectable change on BKL and a negligible one on NV. The AKIEC
and BCC gains have $p<0.05$ at every
threshold and the DF gain at three of four, whereas the VASC and MEL losses reach $p<0.05$ at one
threshold each (VASC at 0.97, MEL at 0.90). Against random sampling the algorithm is higher in every
class at every threshold, most clearly on BCC, DF and VASC. The per-class tests are uncorrected, and
AKIEC, DF and VASC have only 35 to 44 test images each, so each class score is noisy. The direction of
the VASC result matches the tournament, where FID-Inception was the strongest metric on that class
(Section~\ref{sec:tournament-no-single-metric-serves-every-class}). The macro gain over FID-Inception
comes from the other classes outweighing this loss.

\begin{table}[t]
\centering
\caption{Swin-B per-class test $F_2$ difference of the algorithm from the best-metric cohort
(top) and from random sampling (bottom), at each safety threshold $\tau$, with Welch $p$-values
(uncorrected for multiple comparisons). The last row of each block is Macro-$F_2$.}
\label{tab:benchmark-class-diff}
\footnotesize
\setlength{\tabcolsep}{3pt}
\begin{tabular}{@{}lrrrrrrrr@{}}
\toprule
 & \multicolumn{2}{c}{$\tau=0.90$} & \multicolumn{2}{c}{$\tau=0.93$} & \multicolumn{2}{c}{$\tau=0.97$} & \multicolumn{2}{c}{$\tau=0.99$} \\
\cmidrule(lr){2-3}\cmidrule(lr){4-5}\cmidrule(lr){6-7}\cmidrule(lr){8-9}
Class & $\Delta$ & $p$ & $\Delta$ & $p$ & $\Delta$ & $p$ & $\Delta$ & $p$ \\
\midrule
\textit{Algorithm vs.\ Best-metric (FID-Inception)} & & & & & & & & \\
AKIEC & +0.061 & $2.6\times10^{-4}$ & +0.045 & 0.0074 & +0.043 & 0.0066 & +0.040 & 0.0128 \\
BCC & +0.056 & $2.8\times10^{-4}$ & +0.062 & $1.2\times10^{-4}$ & +0.066 & $9.8\times10^{-5}$ & +0.047 & 0.0042 \\
BKL & +0.009 & 0.2837 & +0.004 & 0.6813 & $-$0.001 & 0.9401 & +0.001 & 0.9190 \\
DF & +0.026 & 0.0253 & +0.020 & 0.0750 & +0.031 & 0.0054 & +0.038 & $8.8\times10^{-4}$ \\
MEL & $-$0.027 & 0.0237 & $-$0.007 & 0.6187 & $-$0.013 & 0.2418 & $-$0.010 & 0.3678 \\
NV & +0.003 & 0.0458 & +0.002 & 0.2360 & +0.001 & 0.4256 & +0.002 & 0.1940 \\
VASC & $-$0.019 & 0.1083 & $-$0.011 & 0.3523 & $-$0.031 & 0.0060 & $-$0.020 & 0.0897 \\
Macro & +0.016 & 0.0016 & +0.016 & 0.0016 & +0.014 & 0.0039 & +0.014 & 0.0024 \\
\midrule
\textit{Algorithm vs.\ Random} & & & & & & & & \\
AKIEC & +0.030 & 0.1137 & +0.014 & 0.4592 & +0.012 & 0.5164 & +0.009 & 0.6252 \\
BCC & +0.051 & $6.3\times10^{-4}$ & +0.056 & $2.7\times10^{-4}$ & +0.061 & $2.1\times10^{-4}$ & +0.042 & 0.0087 \\
BKL & +0.017 & 0.0343 & +0.011 & 0.1688 & +0.007 & 0.3772 & +0.009 & 0.2723 \\
DF & +0.045 & $1.8\times10^{-4}$ & +0.039 & $8.4\times10^{-4}$ & +0.050 & $1.7\times10^{-5}$ & +0.057 & $2.0\times10^{-6}$ \\
MEL & +0.014 & 0.2395 & +0.035 & 0.0124 & +0.028 & 0.0224 & +0.032 & 0.0065 \\
NV & +0.004 & 0.0047 & +0.003 & 0.0306 & +0.002 & 0.0839 & +0.003 & 0.0190 \\
VASC & +0.049 & $8.7\times10^{-4}$ & +0.058 & $1.2\times10^{-4}$ & +0.037 & 0.0087 & +0.048 & 0.0011 \\
Macro & +0.030 & $1.5\times10^{-7}$ & +0.031 & $2.2\times10^{-7}$ & +0.028 & $3.7\times10^{-7}$ & +0.029 & $2.0\times10^{-7}$ \\
\bottomrule

\end{tabular}
\end{table}

\begin{table}[t]
\centering
\caption{Swin-B per-class test $F_2$ (mean over 30 seeds; the real-only row is one run). The last
column is Macro-$F_2$. $\tau$ is the safety threshold. Test images per class: AKIEC 43, BCC 93, BKL 217, DF 44, MEL 171, NV 909,
VASC 35.}
\label{tab:benchmark-per-class}
\small
\setlength{\tabcolsep}{4pt}
\begin{tabular}{@{}lrrrrrrrr@{}}
\toprule
Cohort & AKIEC & BCC & BKL & DF & MEL & NV & VASC & Macro \\
\midrule
Real data only & 0.421 & 0.469 & 0.530 & 0.452 & 0.457 & 0.914 & 0.364 & 0.515 \\
\midrule
Random ($N=8{,}012$) & 0.458 & 0.520 & 0.577 & 0.457 & 0.481 & 0.909 & 0.487 & 0.556 \\
Best-metric (FID-Inception) ($N=8{,}012$) & 0.427 & 0.515 & 0.585 & 0.476 & 0.522 & 0.910 & 0.555 & 0.570 \\
\midrule
Algorithm, $\tau=0.90$ ($N=8{,}012$) & 0.488 & 0.572 & 0.594 & 0.502 & 0.495 & 0.913 & 0.536 & 0.586 \\
Algorithm, $\tau=0.93$ ($N=8{,}012$) & 0.472 & 0.577 & 0.589 & 0.496 & 0.516 & 0.912 & 0.545 & 0.587 \\
Algorithm, $\tau=0.97$ ($N=8{,}012$) & 0.470 & 0.581 & 0.584 & 0.507 & 0.509 & 0.911 & 0.524 & 0.584 \\
Algorithm, $\tau=0.99$ ($N=8{,}012$) & 0.467 & 0.562 & 0.586 & 0.514 & 0.513 & 0.912 & 0.535 & 0.584 \\
\midrule
Random ($N=20{,}000$) & 0.496 & 0.540 & 0.545 & 0.420 & 0.482 & 0.910 & 0.494 & 0.555 \\
Algorithm, $\tau=0.97$ ($N=20{,}000$) & 0.526 & 0.601 & 0.571 & 0.482 & 0.495 & 0.913 & 0.511 & 0.586 \\
\bottomrule

\end{tabular}
\end{table}

Table~\ref{tab:tail-stats} reports the tail of the nearest-real similarity for each Swin-B cohort,
because FSMS is a mean and does not show the closest images. The cap works as designed: at $\tau=0.90$
and $0.93$ the maximum equals the threshold, and at $\tau=0.97$ no image reaches 0.97. The algorithm at
$\tau=0.97$ and $0.99$ keeps about 1.8\% of its images at or above 0.95, against 0.43\% for random
sampling, and at $\tau=0.99$ the worst seed contains the same closest pool image as random sampling
and the best-metric cohort (0.9807). The ResNet-50 cohorts at $\tau=0.97$ show the same pattern (mean
maximum 0.9681, 132.3 images at or above 0.95, none at or above 0.97).

\begin{table}[t]
\centering
\caption{Tail of the nearest-real DINOv2 cosine similarity of the images in each Swin-B cohort
($N=8{,}012$ per set, 30 seeds). Max is the largest similarity in a set, given as the mean over
seeds and as the value of the worst seed. p99.9 is the 99.9th percentile of a set, averaged over
seeds. The last two columns give the mean number (share) of images at or above 0.95 and 0.97.}
\label{tab:tail-stats}
\small
\setlength{\tabcolsep}{4pt}
\begin{tabular}{@{}lrrrrr@{}}
\toprule
Cohort & Max (mean) & Max (worst seed) & p99.9 & $\geq 0.95$ & $\geq 0.97$ \\
\midrule
Random & 0.9674 & 0.9807 & 0.9566 & 34.2 (0.43\%) & 0.3 (0.004\%) \\
Best-metric (FID-Inception) & 0.9790 & 0.9807 & 0.9638 & 149.1 (1.86\%) & 2.5 (0.032\%) \\
\midrule
Algorithm ($\tau=0.90$) & 0.9000 & 0.9000 & 0.8999 & 0.0 (0.00\%) & 0.0 (0.000\%) \\
Algorithm ($\tau=0.93$) & 0.9300 & 0.9300 & 0.9299 & 0.0 (0.00\%) & 0.0 (0.000\%) \\
Algorithm ($\tau=0.97$) & 0.9686 & 0.9698 & 0.9627 & 141.7 (1.77\%) & 0.0 (0.000\%) \\
Algorithm ($\tau=0.99$) & 0.9735 & 0.9807 & 0.9631 & 141.0 (1.76\%) & 1.4 (0.017\%) \\
\bottomrule

\end{tabular}
\end{table}

\section{ResNet-50 Results in Full}
\label{app:resnet}

Table~\ref{tab:resnet-paired} gives the difference between each distributional arm and the random control
in the ResNet-50 tournament, Table~\ref{tab:resnet-leaders} the best arm per class on each architecture, and
Table~\ref{tab:resnet-weights} the comparison of the weights. Table~\ref{tab:resnet-temperature} gives the
temperature sweep. Table~\ref{tab:cohorts-resnet} gives the benchmark cohorts,
Table~\ref{tab:benchmark-stats-resnet} their comparisons, and
Table~\ref{tab:benchmark-per-class-resnet} the per-class test $F_2$ for ResNet-50, the counterpart of
Table~\ref{tab:benchmark-per-class}.

\begin{table}[t]
\centering
\caption{Difference in ResNet-50 Macro-$F_2$ between each distributional arm and the random control
(30 seeds each), with the $p$-value of Welch's $t$-test.}
\label{tab:resnet-paired}
\small
\begin{tabular}{@{}lrr@{}}
\toprule
Arm & $\Delta$ & $p$ \\
\midrule
FID-Inception & +0.041 & $2.2\times10^{-7}$ \\
SWD-DINOv2 & +0.035 & $2.2\times10^{-5}$ \\
FID-DINOv2 & +0.033 & $7.5\times10^{-5}$ \\
KID-DINOv2 & +0.033 & $7.5\times10^{-5}$ \\
\bottomrule

\end{tabular}
\end{table}

\begin{table}[t]
\centering
\caption{Best arm per class on the validation split, with its $F_2$, for each architecture, and the
Spearman correlation of the arms' per-class $F_2$ between the architectures (15 distinct arms).}
\label{tab:resnet-leaders}
\small
\begin{tabular}{@{}llll@{}}
\toprule
Class & Swin-B leader & ResNet-50 leader & Rank correlation \\
\midrule
AKIEC & FSIM (0.596) & FSIM (0.599) & 0.69 \\
BCC & Realism (0.597) & FID-Inception (0.573) & 0.73 \\
BKL & Realism (0.561) & IS contribution (0.533) & 0.37 \\
DF & FS-CS (0.225) & CLIP log-odds (0.300) & 0.50 \\
MEL & Realism (0.445) & PSNR (0.397) & 0.51 \\
NV & FID-DINOv2/KID-DINOv2 (0.936) & Diversity (0.929) & 0.69 \\
VASC & FID-Inception (0.680) & Realism (0.625) & 0.74 \\
\bottomrule

\end{tabular}
\end{table}

\begin{table}[t]
\centering
\caption{Distributional share of the weight and largest single weight (with its metric) per class,
for each architecture.}
\label{tab:resnet-weights}
\small
\setlength{\tabcolsep}{4pt}
\begin{tabular}{@{}lrlrl@{}}
\toprule
 & \multicolumn{2}{c}{Swin-B} & \multicolumn{2}{c}{ResNet-50} \\
\cmidrule(lr){2-3}\cmidrule(lr){4-5}
Class & Dist.\ share & Top metric & Dist.\ share & Top metric \\
\midrule
AKIEC & 22\% & FSIM (13\%) & 26\% & KID-DINOv2 (26\%) \\
BCC & 28\% & FID-Inception (13\%) & 31\% & FID-Inception (12\%) \\
BKL & 24\% & SWD-DINOv2 (15\%) & 28\% & KID-DINOv2 (18\%) \\
DF & 33\% & SWD-DINOv2 (33\%) & 21\% & CLIP log-odds (22\%) \\
MEL & 23\% & FID-Inception (13\%) & 31\% & FID-Inception (19\%) \\
NV & 25\% & KID-DINOv2 (13\%) & 24\% & Diversity (9\%) \\
VASC & 40\% & FID-Inception (35\%) & 28\% & FID-Inception (15\%) \\
\bottomrule

\end{tabular}
\end{table}

\begin{table}[t]
\centering
\caption{Validation Macro-$F_2$ of the algorithm on ResNet-50 at seven temperatures (locked configuration
with safety threshold $\tau=0.97$, mean $\pm$ standard deviation over 30 curation seeds). $\Delta$ is the
difference from $T=0.05$ and $p$ is Welch's $t$-test against it.}
\label{tab:resnet-temperature}
\small
\begin{tabular}{@{}lccr@{}}
\toprule
$T$ & Macro-$F_2$ & $\Delta$ vs.\ $T{=}0.05$ & $p$ \\
\midrule
0.02 & $0.4776 \pm 0.0316$ & $-$0.0278 & $4.0\times10^{-4}$ \\
0.05 & $0.5054 \pm 0.0253$ & --- & --- \\
0.1 & $0.4913 \pm 0.0220$ & $-$0.0141 & 0.0249 \\
0.3 & $0.4658 \pm 0.0214$ & $-$0.0396 & $<10^{-6}$ \\
0.5 & $0.4607 \pm 0.0241$ & $-$0.0448 & $<10^{-6}$ \\
1.0 & $0.4663 \pm 0.0283$ & $-$0.0391 & $<10^{-6}$ \\
2.0 & $0.4535 \pm 0.0232$ & $-$0.0519 & $<10^{-6}$ \\
\bottomrule

\end{tabular}
\end{table}

\begin{table}[t]
\centering
\caption{ResNet-50 test-set results. Entries are mean $\pm$ standard deviation over 30 curation
seeds, and the real-only row is a single run.}
\label{tab:cohorts-resnet}
\small
\begin{tabular}{@{}lrrcc@{}}
\toprule
Cohort & Synthetic $N$ & Runs & Macro-$F_2$ & FSMS \\
\midrule
Real data only & --- & 1 & 0.5578 & --- \\
\midrule
Random & $8{,}012$ & 30 & 0.5466 $\pm$ 0.0203 & 0.1639 $\pm$ 0.0009 \\
Best-metric (FID-Inception) & $8{,}012$ & 30 & 0.5577 $\pm$ 0.0250 & 0.1157 $\pm$ 0.0003 \\
\midrule
Algorithm, $\tau=0.97$ & $8{,}012$ & 30 & 0.5520 $\pm$ 0.0222 & 0.1165 $\pm$ 0.0004 \\
\midrule
Random & $20{,}000$ & 30 & 0.5560 $\pm$ 0.0241 & 0.1638 $\pm$ 0.0006 \\
Algorithm, $\tau=0.97$ & $20{,}000$ & 30 & 0.5536 $\pm$ 0.0214 & 0.1183 $\pm$ 0.0003 \\
\bottomrule

\end{tabular}
\end{table}

\begin{table}[t]
\centering
\caption{ResNet-50 test-set comparisons of Macro-$F_2$ (difference, 95\% confidence interval from
Welch's $t$-test, Cohen's $d$, Welch $p$). The algorithm is run at $\tau=0.97$.}
\label{tab:benchmark-stats-resnet}
\footnotesize
\setlength{\tabcolsep}{3pt}
\begin{tabular}{@{}lrrrr@{}}
\toprule
Comparison & $\Delta$ & 95\% CI & $d$ & Welch $p$ \\
\midrule
Algorithm vs.\ Best-metric, $N=8{,}012$ & $-$0.0058 & [$-$0.0180, +0.0064] & $-$0.24 & 0.3481 \\
Algorithm vs.\ Random, $N=8{,}012$ & +0.0054 & [$-$0.0056, +0.0164] & 0.25 & 0.3306 \\
Best-metric vs.\ Random, $N=8{,}012$ & +0.0112 & [$-$0.0006, +0.0229] & 0.49 & 0.0631 \\
Algorithm vs.\ Random, $N=20{,}000$ & $-$0.0024 & [$-$0.0141, +0.0094] & $-$0.10 & 0.6871 \\
Algorithm, $N=20{,}000$ vs.\ $N=8{,}012$ & +0.0017 & [$-$0.0096, +0.0129] & 0.08 & 0.7647 \\
\bottomrule

\end{tabular}
\end{table}

\begin{table}[t]
\centering
\caption{ResNet-50 per-class test $F_2$ (mean over 30 seeds; the real-only row is one run). The last
column is Macro-$F_2$. Test images per class: AKIEC 43, BCC 93, BKL 217, DF 44, MEL 171, NV 909,
VASC 35.}
\label{tab:benchmark-per-class-resnet}
\small
\setlength{\tabcolsep}{4pt}
\begin{tabular}{@{}lrrrrrrrr@{}}
\toprule
Cohort & AKIEC & BCC & BKL & DF & MEL & NV & VASC & Macro \\
\midrule
Real data only & 0.518 & 0.663 & 0.646 & 0.368 & 0.388 & 0.903 & 0.419 & 0.558 \\
\midrule
Random ($N=8{,}012$) & 0.490 & 0.550 & 0.619 & 0.480 & 0.410 & 0.916 & 0.362 & 0.547 \\
Best-metric (FID-Inception) ($N=8{,}012$) & 0.472 & 0.561 & 0.636 & 0.486 & 0.426 & 0.913 & 0.410 & 0.558 \\
\midrule
Algorithm, $\tau=0.97$ ($N=8{,}012$) & 0.484 & 0.580 & 0.615 & 0.481 & 0.417 & 0.916 & 0.371 & 0.552 \\
\midrule
Random ($N=20{,}000$) & 0.567 & 0.508 & 0.600 & 0.509 & 0.412 & 0.914 & 0.384 & 0.556 \\
Algorithm, $\tau=0.97$ ($N=20{,}000$) & 0.547 & 0.556 & 0.602 & 0.482 & 0.398 & 0.913 & 0.378 & 0.554 \\
\bottomrule

\end{tabular}
\end{table}

\section{Metric Computation Details}
\label{app:metrics}

This section specifies how each score of Table~\ref{tab:portfolio} is computed, in the order of
Figure~\ref{fig:metric-pipeline}. All scores are computed once for the whole pool and stored in one table,
and the tournament selects from that table.

\paragraph{Standardization and features.}
Real images are center-cropped to a square on the shorter side and resized to $512 \times 512$ with
Lanczos resampling, the same procedure as for the generator's training data. Synthetic images are
generated at $512 \times 512$. DINOv2 features are the CLS token of \texttt{facebook/dinov2-base} (768
dimensions) obtained with the model's default image processor. InceptionV3 uses the torchvision
\texttt{IMAGENET1K\_V1} weights on images resized to $299 \times 299$ and ImageNet-normalized, and
yields the 2,048-dimensional average-pooled features and the softmax over the 1,000 ImageNet classes.

\paragraph{Nearest-neighbor search.}
All neighbor searches are exact (a flat index in FAISS \citep{Johnson2017faiss}, on GPU when
available) on DINOv2 features that are L2-normalized, so the Euclidean distance is a monotone function
of the cosine similarity. The reference for the full-reference metrics is the nearest of all 8,012 real
training images to each synthetic image, regardless of class.

\paragraph{Full-reference metrics.}
For each synthetic image and its nearest real image, PSNR, SSIM and FSIM are computed after both images
are resized to $256 \times 256$ (with anti-aliasing). PSNR is $20\log_{10}(255/\sqrt{\text{MSE}})$ over
the RGB values, and it is set to 100 when the images are identical. SSIM uses the scikit-image
implementation with its default window, over the color channels with unit data range. FSIM is computed
on the luminance channel from phase congruency (the \texttt{phasepack} implementation) and Sobel gradient
magnitude, with similarity constants $T_1 = 0.085$ and $T_2 = 160$ and equal exponent on the two maps,
and it is weighted by the larger phase congruency of the pair. LPIPS uses the AlexNet backbone of the
\texttt{lpips} package on the images resized to $256 \times 256$ and scaled to $[-1, 1]$.

\paragraph{No-reference metric.}
BRISQUE is computed on each full image with the \texttt{brisque} Python package. Lower is better.

\paragraph{Semantic metrics.}
FS-CS is the cosine similarity between an image's DINOv2 feature and the mean DINOv2 feature of the
real training images of the class the image was generated for. The CLIP scores use
\texttt{openai/clip-vit-base-patch32} and two prompts, ``A clinical photograph of a skin cancer
lesion'' and ``A computer generated image, an artificial image, a fake image''. The two similarities
are the image--text logits of the model, and the log-odds score is $\log\bigl(s_{\text{cancer}} /
(s_{\text{artifact}} + \epsilon)\bigr)$ with a small $\epsilon$. The IS contribution of an
image is $D_{\mathrm{KL}}\bigl(p(y\mid x)\,\|\,p(y)\bigr)$, where $p(y\mid x)$ is the InceptionV3
softmax and $p(y)$ is the mean of $p(y\mid x)$ over the whole pool.

\paragraph{Geometric metrics.}
Realism is the Euclidean distance between an image and its 3rd nearest real training image, and Diversity
is the distance to its 3rd nearest other synthetic image in the whole pool, both on L2-normalized DINOv2
features. The tournament selects the lowest Realism distance and the highest Diversity distance.

\paragraph{Distributional metrics.}
Each of the 16 configurations (17,217 images) is compared with the 8,012 real training images. FID is the
Fr\'echet distance between Gaussians fitted to the full sets, on the 2,048-dimensional InceptionV3
features (FID-Inception) or the 768-dimensional DINOv2 features (FID-DINOv2). KID and SWD use the DINOv2
features of equal-sized random subsamples of 8,012 images from each set, drawn with a fixed seed. KID is
the squared maximum mean discrepancy with the cubic polynomial kernel $k(x, y) = (x^\top y / d + 1)^3$,
estimated from the full kernel matrices including the diagonal terms. SWD is the mean absolute difference
between the sorted projections onto 128 random unit directions. Lower is better for all four.
The group scores of Section~\ref{sec:algorithm-group-scored-distributional-te} use the same statistics on the
full group, except that SWD uses a random sample of the group matched in size to the real set (8,012 images).

\section{Reproducibility Details}
\label{app:repro}

\paragraph{Data.}
The real training set (8{,}012 images) and validation set (2{,}003 images) are a class-stratified
80/20 re-partition of the 10{,}015-image ISIC-2018 training archive, and the test set is the
official ISIC-2018 Task 3 release (1{,}512 images). All images are center-cropped and resized to
$512 \times 512$ before feature extraction (Appendix~\ref{app:metrics}).

\paragraph{Classifiers.}
Torchvision Swin-B and ResNet-50 with ImageNet (\texttt{IMAGENET1K\_V1}) weights, backbone frozen,
linear head trained with AdamW (learning rate $10^{-3}$), cross-entropy loss, batch size 32, at most
50 epochs, early stopping with patience 10 on validation accuracy. Training images use random
resized crops to 224 pixels and horizontal flips. Validation and test images are resized to 256
and center-cropped to 224. Weight initialization and data-loader seed are fixed at 42, with cuDNN
deterministic mode on and benchmark mode off.

\paragraph{Seeds.}
Distributional tournament arms use curation seeds 42 to 71, and the random control uses the same
seeds, extended to 42 to 101 for Swin-B. All cohorts of the benchmark use the curation seeds 100 to 129. The
algorithm cohorts share these seeds across the four safety thresholds, and the cohorts draw from
different pools, so the same seed number gives unrelated draws.

\paragraph{Compute.}
EDM2 training ran on three NVIDIA L40 GPUs, and pool generation on one L40 per array task, on the
Bunya cluster \citep{rcc2024bunya}. Classifier training for the tournament, the algorithm development and the
benchmark ran on a four-GPU NVIDIA L40S server without a job scheduler. Software versions of the
classification environment are PyTorch 2.2.0, torchvision 0.17.0, scikit-learn 1.4.1 and pandas 2.2.1
(Python 3.10).

\end{document}